\documentclass[10pt,twocolumn,letterpaper]{article}

\usepackage[pagenumbers]{wacv}

\usepackage[utf8]{inputenc}
\usepackage[T1]{fontenc}
\usepackage{amsmath,amssymb}
\usepackage{algorithm}
\usepackage{algpseudocode}
\usepackage{booktabs}
\usepackage{graphicx}
\usepackage{multirow}
\usepackage{float}
\usepackage{placeins}
\usepackage{microtype}
\usepackage{xurl}

\usepackage{amsmath,amsfonts,bm}

\def\eqref#1{Eq.~\ref{#1}}

\def\plaineqref#1{\ref{#1}}

\def\1{\bm{1}}

\DeclareMathAlphabet{\mathsfit}{\encodingdefault}{\sfdefault}{m}{sl}
\SetMathAlphabet{\mathsfit}{bold}{\encodingdefault}{\sfdefault}{bx}{n}

\definecolor{yyyreviewyellow}{RGB}{190,135,0}

\newtheorem{proposition}{Proposition}

\definecolor{wacvblue}{rgb}{0.21,0.49,0.74}
\usepackage[pagebackref,breaklinks,colorlinks,allcolors=wacvblue]{hyperref}

\def\wacvPaperID{}
\def\confName{WACV}
\def\confYear{2027}

\newcommand{\suppTableSetup}{%
  \scriptsize
  \renewcommand{\arraystretch}{1.10}%
  \setlength{\tabcolsep}{2.0pt}%
}

\title{CANVAS: Consistency-Aware Navigation via Visual Adaptive Sampling for Long-Context Text-to-SVG Generation}

\author{%
  Yichen Wu\textsuperscript{1} \quad
  Haoxuan Qu\textsuperscript{2} \quad
  Yihang Lou\textsuperscript{3} \quad
  Hossein Rahmani\textsuperscript{2} \quad
  Jun Liu\textsuperscript{2}\\
  \textsuperscript{1}University of Nottingham \quad
  \textsuperscript{2}Lancaster University \quad
  \textsuperscript{3}Peking University\\[-1pt]
  {\footnotesize Code: \href{https://github.com/Louis-YW/CANVAS}{\texttt{github.com/Louis-YW/CANVAS}}}
}

\begin{document}

\maketitle

\begin{abstract}
Autoregressive large models have recently advanced Text-to-SVG generation from simple icons to complex, long-context graphics, yet standard autoregressive decoding often fails to maintain global consistency across geometry, layout, occlusion, and composition. We introduce \textbf{CANVAS} (\textbf{C}onsistency-\textbf{A}ware \textbf{N}avigation via \textbf{V}isual \textbf{A}daptive \textbf{S}ampling), a training-free, render-aware inference framework that combines power-sharpened trajectory likelihood with visual feedback from rendered futures and derives a stroke-wise navigation rule. It effectively estimates each candidate stroke's future value under a limited generation and rendering budget and adaptively allocates samples according to candidate uncertainty, decision influence, and rollout cost. Experiments across multiple autoregressive SVG backbones and complementary benchmarks demonstrate improvements in global consistency, which includes sound geometric relationships, spatial layouts, occlusion ordering, and overall composition, without additional training, demonstrating the effectiveness and generalization ability of our framework.
\end{abstract}

\section{Introduction}
\label{sec:introduction}

Text-to-SVG generation aims to generate scalable, structurally editable vector graphics from natural-language descriptions, has broad applications in icon design, digital illustration, and web content creation, and has thus attracted increasing research attention~\citep{wu2023iconshop,chen2025svgbuilder,wu2025chat2svg,yang2025omnisvg,rodriguez2025starvector,xing2025llm4svg,wang2025svgen}.
In recent years, with advances in multimodal large language models in code generation and visual understanding, the direct autoregressive generation of SVG code with large models, as a mainstream technical route for this task, has expanded from simple icons to complex graphics containing numerous paths, coordinates, and style attributes~\citep{rodriguez2025starvector,xing2025llm4svg,yang2025omnisvg,zini2025vhector}.

However, the ability to generate longer SVG sequences does not mean that a model can stably produce complete images with good global consistency, which includes sound geometric relationships, spatial layouts, occlusion ordering, and overall composition, all of which are important to the generation quality of complex SVGs~\citep{zini2025vhector,chen2025svgbuilder,yang2025omnisvg,xing2025llm4svg,wu2025chat2svg,wang2025svgen,wang2026introsvg,xing2026reasonsvg,liang2026render}.
This is because achieving these properties depends on a large number of interdependent drawing decisions.
However, standard autoregressive decoding selects only according to the conditional probability at the current position. Therefore, a locally high-probability decision may still lead to a low-quality complete SVG, whereas a globally better generation trajectory may be eliminated early because its early local probabilities are lower and cannot be recovered during subsequent generation.

Such problems do not necessarily mean that the base model has not learned the structure of long-sequence SVGs.
Models trained on large-scale SVG data already contain structural priors about path combinations, stylistic relationships, and complete programs, but standard autoregressive decoding may not fully reflect the model's preference over complete generation trajectories.
Therefore, we study how to use these existing capabilities more effectively at inference time, without modifying the model parameters, to improve the consistency of long-context SVG generation.

However, relying only on the base model's token probabilities remains insufficient, because they mainly reflect the code distribution and generation preferences learned from the training data rather than an explicit evaluation of the final rendering, and thus cannot directly determine whether the shapes, positions, occlusion relationships, and overall composition in the final rendering are consistent.
Even if an SVG trajectory has high model probability and can be parsed successfully, its rendering may still exhibit element misalignment, occlusion conflicts, or compositional imbalance.
Motivated by prior studies that incorporate rendered visual feedback into SVG generation~\citep{liang2026render,wang2026introsvg}, we retain the structural priors learned by the base model while leveraging rendered outputs to provide direct visual feedback.

To this end, we propose CANVAS (Consistency-Aware Navigation via Visual Adaptive Sampling), a training-free, render-aware decoding framework for long-context Text-to-SVG generation.
Here, ``navigation'' refers to selecting generation trajectories in the SVG program space rather than following locally probable token decisions alone.
At the complete-sequence level, CANVAS constructs a unified objective that combines the base model's preference over complete SVG trajectories with the visual quality of the final rendering.
Inspired by professional artists, who consider the global composition when deciding how to place each stroke, we reformulate the complete-sequence objective as a stroke-wise future-aware navigation rule: at each complete drawing boundary identified by an incremental parser, CANVAS navigates among candidate strokes (i.e., newly appended SVG drawing units that are parser-complete and directly renderable) according to their selection probabilities and the overall quality of the future complete SVGs they may produce.

Practical SVG applications also demand efficient inference, as users often need to generate, compare, and modify multiple candidate SVGs~\citep{chen2025svgbuilder,zini2025vhector}.
However, exactly computing the future value of each stroke requires enumerating a large number of complete SVG continuation trajectories, making the associated computational cost prohibitive.
To make this navigation computationally practical, CANVAS uses finite-sample rollouts to obtain conditionally unbiased estimates of finite-horizon future values and adaptively allocates generation and rendering budgets according to the uncertainty, decision influence, and sampling cost of different candidate strokes.
Under theoretical guarantees, this improves the reliability of the estimates under a limited computational budget.

\begin{figure*}[t]
\centering
\includegraphics[width=0.92\textwidth]{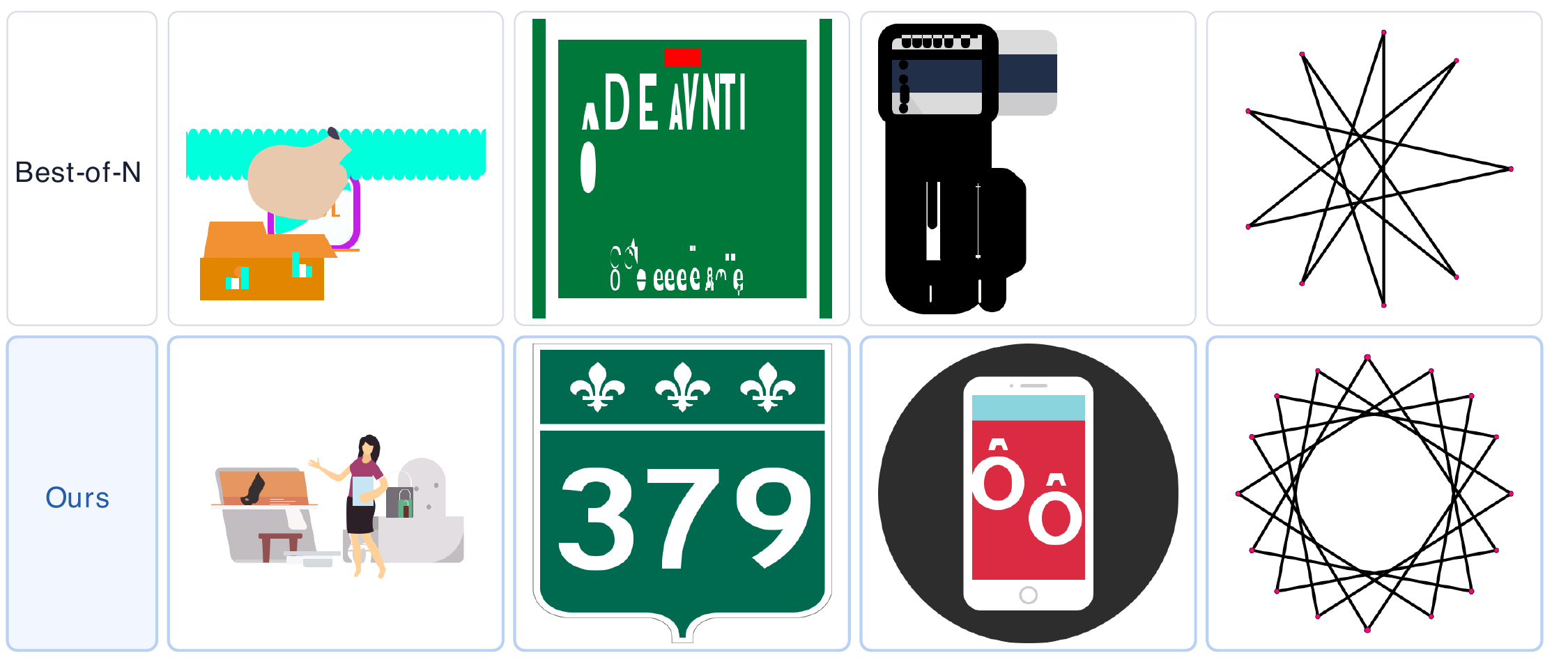}
\caption{Illustration of SVGs generated with the same vHector-8B backbone by Best-of-$N$ and CANVAS at inference time. CANVAS shows better global consistency, which includes sound geometric relationships, spatial layouts, occlusion ordering, and overall composition. The prompts from left to right are: ``The image depicts a cartoon-style illustration featuring a person and a laboratory setup.''; ``The image depicts a traffic sign with a green background and white text and symbols.''; ``The image depicts a mobile phone with a sleek, modern design.''; and ``The image depicts a highly intricate and symmetrical geometric pattern, specifically a star polygon, which is a polygon with multiple vertices and edges that are connected in a star-like shape.''}
\label{fig:qualitative_comparison}
\vspace{-2mm}
\end{figure*}

Figure~\ref{fig:qualitative_comparison} compares CANVAS with Best-of-$N$ decoding.
Even after the unchanged backbone is sampled $N=5$ times and the best completed SVG is selected, CANVAS still exhibits better global consistency, which includes sound geometric relationships, spatial layouts, occlusion ordering, and overall composition.

In summary, our main contributions are as follows: 1) We propose CANVAS, a training-free, render-aware inference framework for consistency-aware navigation in long-context Text-to-SVG generation. 2) We transform the joint model-and-visual objective at the complete-SVG level into a stroke-wise future-aware navigation rule and propose a reliable visual adaptive sampling method with a finite horizon and adaptive budget allocation. 3) Experiments on multiple representative autoregressive SVG generators and complementary benchmarks verify the effectiveness and generalization ability of our method.

\section{Related Work}
\label{sec:related_work}

\paragraph{Text-to-SVG generation.}
Earlier work on this task often adopted optimization-based approaches, synthesizing text-guided vector graphics by directly optimizing vector primitives through differentiable rasterization under vision-language or text-to-image diffusion guidance~\citep{frans2022clipdraw,jain2023vectorfusion,xing2023diffsketcher,xing2024svgdreamer}.
With the advent of MLLMs, recent works mainly follow autoregressive paradigm, tokenizing SVG code and predicting the resulting tokens sequentially~\citep{wu2023iconshop,tang2024strokenuwa,chen2025svgbuilder,wu2025chat2svg,xing2025llm4svg,yang2025omnisvg,zini2025vhector,wang2025svgen}.
Earlier autoregressive methods primarily explore representations that make SVG sequences easier to model: IconShop~\citep{wu2023iconshop} sequentializes SVG paths into uniquely decodable tokens, StrokeNUWA~\citep{tang2024strokenuwa} compresses graphics into semantically meaningful stroke tokens, and SVGBuilder~\citep{chen2025svgbuilder} generates colored graphics with a component-based representation.
Vector Grimoire~\citep{cipriano2025grimoire} learns a discrete codebook of vector shapes from raster supervision.
Chat2SVG~\citep{wu2025chat2svg} uses an LLM to produce an SVG template and then refines its geometry through diffusion-guided optimization.

More recent Text-to-SVG methods seek to improve generation of complex graphics and longer output sequences.
They commonly construct large long-context SVG training datasets~\citep{wang2025svgen,zini2025vhector,chen2025svgthinker,xing2026reasonsvg,wang2026introsvg}.
vHector~\citep{zini2025vhector} introduces domain-specific tokens. IntroSVG~\citep{wang2026introsvg} trains a unified generator and critic and performs an iterative generate, critique, and refine loop using rendered feedback.
Some also introduce Chain-of-Thought (CoT) in supervision~\citep{wang2025svgen,chen2025svgthinker,xing2026reasonsvg}.
SVGen~\citep{wang2025svgen} combines curriculum learning with integrity and path-count rewards, while SVGThinker~\citep{chen2025svgthinker} aligns CoT supervision with incremental rendering, and Reason-SVG~\citep{xing2026reasonsvg} introduces Drawing-with-Thought with hybrid-reward reinforcement learning.

Together, these methods modify the training data, representation, supervision, or inference loop.

Different from existing studies, to the best of our knowledge, CANVAS is the first training-free inference framework to improve the global consistency of SVGs generated by diverse autoregressive backbones.

\paragraph{Power-distribution sampling.}
Power-distribution sampling globally sharpens a model's sequence distribution, allowing inference to favor trajectories that are probable as complete sequences rather than relying only on local token probabilities.
Karan and Du~\citep{karan2025reasoning} show that this principle can elicit strong reasoning from base LLMs without additional training, and it has subsequently been extended to long-horizon robotic planning~\citep{chen2026drift}.
Sampling for Quality~\citep{markovicvoronov2026sampling} augments the base sequence distribution with prefix-dependent reward potentials, enabling model likelihood and sequence-level quality to jointly guide decoding.
Different from existing works, CANVAS makes decisions at the stroke level by combining each candidate's power-sharpened model probability with the visual quality of the future complete SVGs it may lead to.

\section{Method}
\label{sec:method}

\begin{figure*}[t]
\centering
\includegraphics[width=0.92\textwidth]{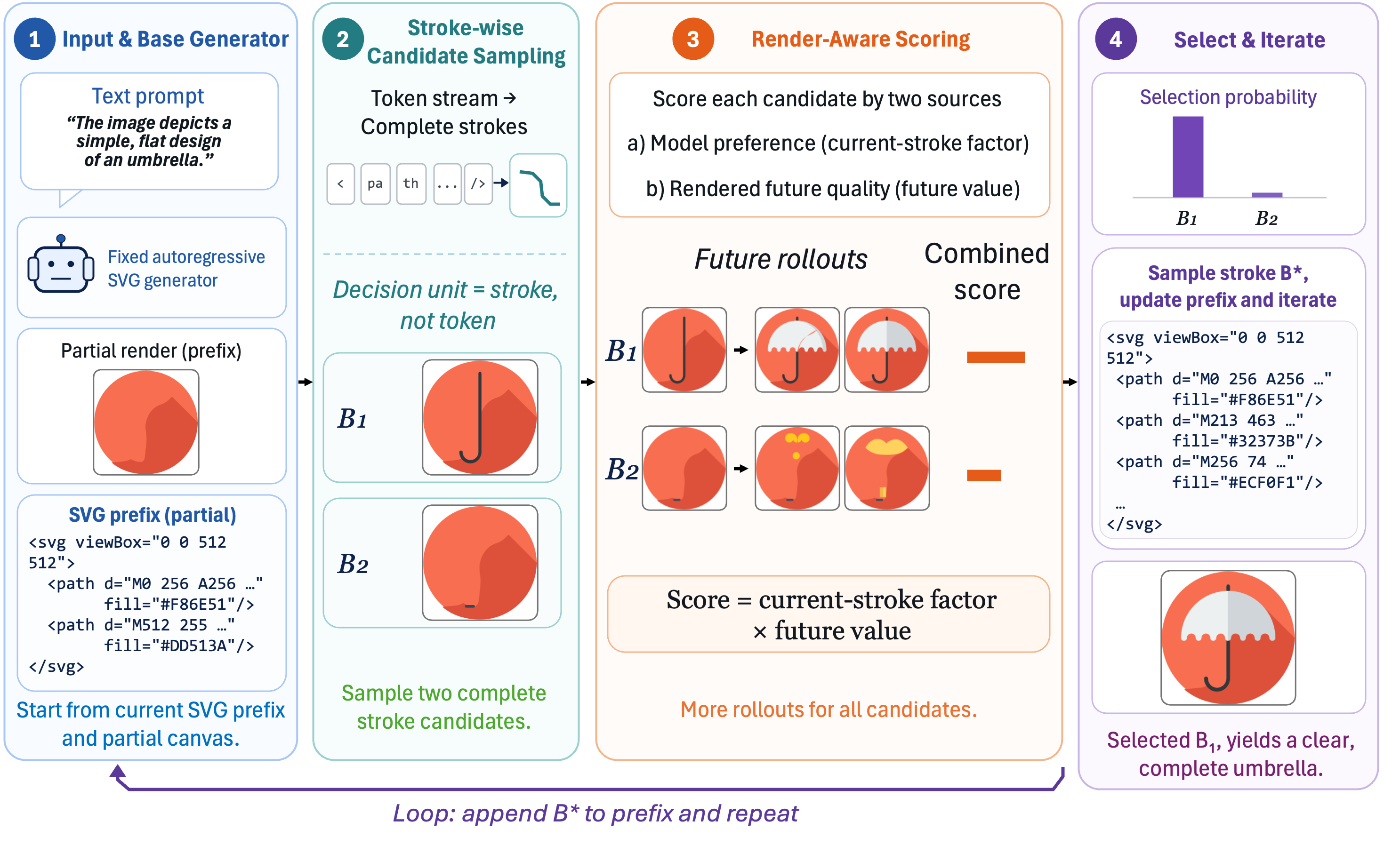}
\caption{\textbf{Overview of CANVAS.} Long-context SVG decoding is difficult because a locally plausible stroke can lead to globally inconsistent geometry, occlusion, and composition, while exhaustively scoring all complete futures is prohibitive in generation and rendering costs. CANVAS addresses this challenge with a novel training-free, render-aware stroke-wise navigation mechanism: at each parser-complete boundary, it samples candidate strokes, evaluates their future rendered trajectories with a joint model-and-visual score, adaptively concentrates rollout budget on uncertain and decision-critical candidates, and commits one stroke before iterating with the fixed base generator.}
\label{fig:canvas_overview}
\vspace{-2mm}
\end{figure*}

To improve global consistency in long-context SVG generation during inference time, a key challenge is that standard autoregressive decoding acts only on local token probabilities and cannot anticipate how a current choice will affect the final canvas.
To tackle this challenge, we propose CANVAS, a novel framework that makes stroke-wise decisions by combining power-sharpened model preferences with visual evidence from the rendered futures that each candidate stroke may produce.
Figure~\ref{fig:canvas_overview} provides an overview of CANVAS.

Sec.~\ref{sec:problem_formulation} defines a complete-SVG target that combines trajectory-level model likelihood with rendered visual quality.
Sec.~\ref{sec:stroke_framework} derives its parser-aligned stroke-wise navigation rule, Sec.~\ref{sec:finite_sampling} replaces the intractable candidate and future spaces with finite sampling, and Sec.~\ref{sec:reliable_future_estimation} allocates a limited generation and rendering budget across candidates.

\subsection{Complete-SVG Target}
\label{sec:problem_formulation}

Given a natural-language prompt $q$, the base model generates a complete SVG program $Y=(y_1,\ldots,y_{T(Y)})$ with probability
\begin{equation}
    p_\theta(Y\mid q)
    =\prod_{t=1}^{T(Y)}p_\theta(y_t\mid y_{<t},q).
    \label{eq:token_factorization}
\end{equation}
Inspired by Context-Aware Power Sampling (CAPS)~\citep{chen2026drift}, we first investigate whether power sharpening of this model's joint trajectory likelihood can improve global consistency in long-context SVG generation. However, as shown by the comparison in Appendix~\ref{app:power_ablation}, this model-likelihood-only approach remains insufficient.
We attribute this limitation to the fact that model likelihood cannot determine whether the rendered SVG has correct geometry, occlusion relations, and overall composition.
Direct visual feedback is therefore necessary; maximizing it alone, however, could exploit evaluator errors and abandon the SVG prior.
Effectively combining these two sources of information without discarding either one is the first challenge.
Let $p_{\theta,\alpha}^{\mathrm{pow}}(Y\mid q)\propto p_\theta(Y\mid q)^\alpha$, with $\alpha\geq1$, denote the power-sharpened model distribution.
Inspired by the KL-regularized preference-optimization principle underlying DPO~\citep{rafailov2023dpo}, CANVAS retains the power-sharpened generator as a frozen prior while incorporating rendered visual quality as an inference-time reward.
We seek a visually improved distribution through the following KL-regularized objective~\citep{markovicvoronov2026sampling}:
\begin{equation}
\begin{aligned}
    \Pi_{\alpha,\beta}(\cdot\mid q)
    &=\arg\max_{\pi\in\Delta(\mathcal Y_q)}
    \mathcal J_{\alpha,\beta}(\pi),\\
    \mathcal J_{\alpha,\beta}(\pi)
    &=\beta\,\mathbb E_{Y\sim\pi}[s(\mathcal R(Y),q)]\\[-1mm]
    &\quad-D_{\mathrm{KL}}\!\left(
    \pi(\cdot\mid q)\,\middle\|\,
    p_{\theta,\alpha}^{\mathrm{pow}}(\cdot\mid q)
    \right),
    \qquad \beta\geq0.
\end{aligned}
    \label{eq:main_variational_objective}
\end{equation}
Here, $\mathcal R(Y)$ renders program $Y$, and $s(\mathcal R(Y),q)\in[0,1]$ measures its visual quality relative to prompt $q$.
The first term favors visually coherent complete renderings, whereas the KL term penalizes departure from the model's power-sharpened trajectory distribution.
The objective therefore improves final-render quality while retaining the SVG syntax, structure, and complete-program preferences learned by the base model.
Its unique optimizer has the closed form
\begin{equation}
\begin{aligned}
    \Pi_{\alpha,\beta}(Y\mid q)
    &=\frac{1}{\widetilde Z_{\alpha,\beta}(q)}
    p_\theta(Y\mid q)^\alpha
    \\[-1mm]
    &\quad\times\exp\!\{\beta s(\mathcal R(Y),q)\},
    \qquad Y\in\mathcal Y_q,
\end{aligned}
    \label{eq:complete_svg_target}
\end{equation}
where $\widetilde Z_{\alpha,\beta}(q)$ normalizes the distribution over valid, terminated SVGs $\mathcal Y_q$.
The factor $p_\theta(Y\mid q)^\alpha$ preserves and strengthens the model's preference over complete programs, while $\exp\{\beta s(\mathcal R(Y),q)\}$ reweights them according to final-render quality.
Thus, the target favors complete SVGs that are both plausible to the generator and visually consistent after rendering.
The detailed derivation from the KL-regularized objective to the Gibbs distribution above is provided in Appendix~\ref{app:variational_target}.

\subsection{Consistency-Aware Stroke-wise Navigation}
\label{sec:stroke_framework}

Eq.~\ref{eq:complete_svg_target} specifies how probability mass should be assigned to complete, terminated SVG programs, but it does not directly provide an online decision rule.
During decoding, we must decide which candidate generation unit to choose before the future suffix is generated, evaluating the set of possible complete SVG trajectories that each candidate unit may induce.
Inspired by the global-to-step-wise reformulation of Scalable Power Sampling~\citep{ji2026scalable}, but using a distinct render-aware objective, we first consider token-wise decisions; these require prohibitively frequent rollouts (Appendix~\ref{app:decision_reward_ablation}).
A naive way to reduce this cost is to replace token-wise decoding with fixed-length blocks.
Although this reduces the number of decisions, block boundaries may fall inside a coordinate, path command, or XML attribute and thus fail to represent a complete visual operation.
Inspired by path-level SVG representations and step-wise path-fragment generation~\citep{wu2023iconshop,liang2026render}, CANVAS instead parses the SVG token stream as it is generated and places a decision boundary whenever a complete renderable element (hereafter called a stroke) has been formed.
We operationally define a stroke as a parser-complete SVG drawing unit, either a \texttt{<path>} element or another directly renderable primitive, such as \texttt{<circle>} or \texttt{<rect>}.
More details about the mathematical forms for token-wise, fixed-length, and parser-detected blocks are provided in Appendix~\ref{app:boundary_rule}.

Each stroke-boundary prefix $H_k$ then corresponds to a visually meaningful partial rendering $\mathcal R(H_k)$.
Let $h=H_{k-1}$ be the current SVG prefix, let $b$ be a candidate next stroke, and write $H_k=hb$ after committing it.
To measure only the visual change caused by $b$, rather than repeatedly rewarding the existing canvas and thereby favoring programs with more strokes, we define the incremental render potential
\begin{equation}
    \psi_k(H_k,q)
    =\exp\!\left\{
    \beta\big[s(\mathcal R(H_k),q)-s(\mathcal R(H_{k-1}),q)\big]
    \right\}.
    \label{eq:render_potential}
\end{equation}
Multiplying these incremental factors over all strokes gives
\begin{equation}
    \prod_{k=1}^{K(Y)}\psi_k(H_k,q)
    =\exp\!\left\{
    \beta\big[s(\mathcal R(Y),q)-s(\mathcal R(H_0),q)\big]
    \right\}.
    \label{eq:telescoping_render_potential}
\end{equation}
Each factor measures only the visual change introduced by one stroke. Multiplying them cancels the intermediate canvas scores, leaving only the visual-reward term for the completed SVG in Eq.~\ref{eq:complete_svg_target}, apart from an initial-canvas constant shared by all candidates.

Because CANVAS chooses a candidate stroke based not only on the model's preference for the current drawing operation but also on the future value of the complete SVG trajectories reachable after that stroke, we divide the remaining target weight of a continuation composed of the current stroke $b$ and future suffix $C$ into two parts: one part is the model preference and immediate visual change determined by $b$, and the other is the total value of all complete continuations that $b$ may induce.
Thus, the complete-trajectory target can be converted into the following online stroke-wise navigation rule:
\begin{equation}
    \boxed{
    \Pi_{\alpha,\beta}(B_k=b\mid H_{k-1}=h,q)
    \propto u_k(b)V_k(hb,q)
    }.
    \label{eq:stroke_conditional}
\end{equation}
where:
\begin{equation}
\begin{aligned}
    u_k(b)
    &=P_\theta^{\mathrm{stroke}}(b\mid h,q)^\alpha\psi_k(hb,q),\\
    V_k(hb,q)
    &=\sum_{C:\,hbC\in\mathcal Y_q}
    p_\theta(C\mid hb,q)^\alpha
    \Phi_{>k}(hb,C,q),\\
    \Phi_{>k}(hb,C,q)
    &=\prod_{r=k+1}^{K(hbC)}\psi_r(H_r,q).
\end{aligned}
\label{eq:stroke_factors}
\end{equation}
Here, $P_\theta^{\mathrm{stroke}}(b\mid h,q)$ is the probability of generating candidate stroke $b$ under the current condition, while $p_\theta(C\mid hb,q)$ is the model probability of generating suffix $C$ after $b$.
The current-stroke factor $u_k(b)$ combines the model's preference for the current stroke with its immediate visual change, whereas the future value $V_k(hb,q)$ marginalizes the model preferences and visual effects of all complete continuations after that stroke.
More derivation details are provided in Appendix~\ref{app:exact_conditional}.

\subsection{From Exact Expectations to Finite Sampling}
\label{sec:finite_sampling}

However, evaluating the future value $V_k(hb,q)$ requires considering a large number of possible complete continuations for every candidate stroke.
As shown in Appendix~\ref{app:horizon_ablation}, doing so incurs prohibitive generation and rendering costs.
Practical SVG applications require users to rapidly generate, compare, and modify multiple candidate results~\citep{chen2025svgbuilder,zini2025vhector}, making complete continuation evaluation impractical.
To make the future value estimable, we express it as an expectation under the base-model continuation distribution:
\begin{equation}
\begin{aligned}
    V_k(hb,q)
    &=\sum_{C:\,hbC\in\mathcal Y_q}
    p_\theta(C\mid hb,q)\\[-1mm]
    &\qquad\times
    \left[
    p_\theta(C\mid hb,q)^{\alpha-1}
    \Phi_{>k}(hb,C,q)
    \right]\\
    &=\mathbb E_{C\sim p_\theta(\cdot\mid hb,q)}\!\Bigl[\\[-1mm]
    &\qquad p_\theta(C\mid hb,q)^{\alpha-1}\\
    &\qquad\times\Phi_{>k}(hb,C,q)
    \Bigr].
\end{aligned}
\label{eq:future_expectation}
\end{equation}
This expectation admits an unbiased Monte Carlo estimator.
We first sample candidates from the parser-valid next-stroke proposal $q_k$ and compute their importance factors:
\begin{equation}
\begin{aligned}
    B_i&\overset{\mathrm{i.i.d.}}{\sim}q_k(\cdot\mid h,q),
    \qquad i=1,\ldots,L,\\
    A_i&=\frac{u_k(B_i)}{q_k(B_i\mid h,q)}.
\end{aligned}
\label{eq:candidate_importance}
\end{equation}
Here, $q_k$ is the parser-valid proposal induced by $P_\theta^{\mathrm{stroke}}$, $L$ is the number of sampled candidates, and $A_i$ is the importance weight for candidate $B_i$. The exact first-hit stopping rule is given in Appendix~\ref{app:boundary_rule}. The resulting importance-weighted average is an unbiased estimate of the exact unnormalized stroke mass:
\begin{equation}
\begin{aligned}
    \mathbb E\!\left[\frac1L\sum_{i=1}^{L}A_iV_k(hB_i,q)\right]
    &=\mathbb E_{B\sim q_k}\!\left[
    \frac{u_k(B)V_k(hB,q)}{q_k(B\mid h,q)}
    \right]\\
    &=\sum_{b\in\mathcal B(h)}u_k(b)V_k(hb,q).
\end{aligned}
\label{eq:finite_candidate_mass}
\end{equation}
At this point, candidate-stroke sampling is complete, but estimating the future value of each candidate remains challenging.
As shown in Tabs.~\ref{tab:main_rollout_horizon_ablation}, the time cost grows with the rollout horizon, whereas rolling out every sample to its end (i.e., its EOS token) incurs unacceptable overhead.
For each candidate $B_i$, we therefore draw $M_i$ independent continuations, each containing at most $H$ future strokes.
Let $C_{1:H}^{(m)}$ denote the continuation from rollout $m$, and let $H_{k+r}^{(m)}$ denote the prefix after its first $r$ future strokes.
In rollout $m$, $R_i^{(m)}$ denotes the number of future strokes included in the rollout: it is $H$ if EOS is not reached within the horizon, and a smaller value if EOS is reached earlier.
We define
\begin{equation}
\begin{aligned}
    Z_{i,H}^{(m)}
    &=p_\theta(C_{1:H}^{(m)}\mid hB_i,q)^{\alpha-1}\\[-1mm]
    &\quad\times
    \prod_{r=1}^{\min(H,R_i^{(m)})}\psi_{k+r}(H_{k+r}^{(m)},q),\\
    \widehat V_{i,H}
    &=\frac1{M_i}\sum_{m=1}^{M_i}Z_{i,H}^{(m)},\\
    \widehat\rho_i
    &=\frac{A_i\widehat V_{i,H}}
    {\sum_{j=1}^{L}A_j\widehat V_{j,H}}.
\end{aligned}
\label{eq:normalized_selection}
\end{equation}
Here, $Z_{i,H}^{(m)}$ is the contribution of rollout $m$ for candidate $B_i$, $\widehat V_{i,H}$ is the sample-mean estimate of its finite-horizon future value, and $\widehat\rho_i$ is the normalized selection weight assigned to $B_i$.
Conditioned on $B_i$, $\widehat V_{i,H}$ is an unbiased estimate of the finite-horizon future value when $M_i$ is fixed before the fresh rollouts are observed.
Together with candidate importance sampling, this gives an unbiased estimate of the finite-horizon unnormalized target mass.
More details about rollout-based future-value estimation are provided in Appendix~\ref{app:finite_horizon}, while details about importance approximation over the finite candidate space are provided in Appendix~\ref{app:importance}.

\subsection{Visual Adaptive Sampling under a Limited Budget}
\label{sec:reliable_future_estimation}

At this stage, candidate strokes differ in both the uncertainty of their future values and the cost of estimating them. Assigning the same rollout count to all candidates wastes computation on stable branches and may undersample candidates whose estimates most affect the normalized stroke decision. To tackle this problem, CANVAS allocates the budget according to
rollout uncertainty, decision influence, and per-rollout cost.

To implement this allocation, we first need candidate-specific estimates of future value, rollout variance, and per-rollout generation-and-rendering
cost. A naive strategy would use the same rollout samples both for this diagnosis and to form the final future-value estimate. This creates selection dependence because the same random fluctuation can influence both the allocated rollout count and the final estimate. CANVAS avoids this dependence by first using $m_0$ pilot rollouts per candidate to estimate its future value $\widehat V_i^{(0)}$, rollout variance $\widehat\sigma_i^2$, and per-rollout generation-and-rendering cost $\widehat c_i$. Then it freezes the allocation and evaluates each candidate with independent fresh rollouts. More details about the derivation of the adaptive rollout allocation are given in Appendix~\ref{app:allocation}, and more
implementation details of the CANVAS decoding procedure and its error decomposition are given in Appendix~\ref{app:procedure}.

Let $V_i=V_{i,H}$, $\sigma_i^2=\operatorname{Var}(Z_{i,H}\mid B_i)$, $D=\sum_iA_iV_i$, and $\rho_i=A_iV_i/D$.
A first-order delta analysis of the normalized map gives the decision-error approximation and its optimal continuous allocation:
\begin{equation}
\begin{aligned}
    \mathbb E\|\widehat{\boldsymbol\rho}-\boldsymbol\rho\|_2^2
    &\simeq\sum_{i=1}^{L}\frac{d_i}{M_i},\\
    d_i&=\frac{A_i^2\sigma_i^2}{D^2}
    \|\mathbf e_i-\boldsymbol\rho\|_2^2,\\
    M_i^*
    &=\frac{C_k\sqrt{d_i/c_i}}
    {\sum_{j=1}^{L}\sqrt{d_jc_j}}.
\end{aligned}
\label{eq:decision_error_surrogate}
\end{equation}
Here, $C_k$ is the fresh-rollout budget for the current decision, $c_i$ is candidate $i$'s per-rollout cost, and $\mathbf e_i$ is its coordinate vector.
In practice, we round the continuous allocation to obtain integer rollout counts.

The allocation above minimizes the first-order decision error, but it remains to determine whether adapting the rollout counts provides a real advantage over assigning the same count to every candidate.
Under the same generation-and-rendering budget, the two error surrogates satisfy
\begin{equation}
\begin{aligned}
    \mathcal E_{\mathrm{adaptive}}
    &=\frac{\left(\sum_i\sqrt{c_id_i}\right)^2}{C_k}\\[-1mm]
    &\leq
    \frac{(\sum_i d_i)(\sum_i c_i)}{C_k}
    =\mathcal E_{\mathrm{uniform}}.
\end{aligned}
    \label{eq:main_uniform_adaptive_comparison}
\end{equation}
where the inequality follows from Cauchy--Schwarz and is strict unless $d_i/c_i$ is identical across candidates.
Thus, under the same budget, our future-uncertainty-aware allocation is theoretically guaranteed to be no worse than uniform allocation.

In addition, a further difficulty remains: even when every $\widehat V_{i,H}$ is conditionally unbiased, the normalized decision in \eqref{eq:normalized_selection} divides by a random denominator, and generally $\mathbb E[X/Y]\neq\mathbb E[X]/\mathbb E[Y]$.
Here, to address this normalization error, we use a second-order expansion of the ratio map.
For a bounded candidate statistic $f$, let $Y_i=A_i\widehat V_{i,H}$, $X_i(f)=f(B_i)Y_i$, and let $\overline Y$, $\overline X_f$, $s_Y^2$, and $s_{XY}(f)$ denote their sample means, denominator variance, and numerator--denominator covariance.
Using these observed second-order statistics, CANVAS subtracts the leading ratio-bias term:
\begin{equation}
\begin{aligned}
    \widetilde T_k(f)
    &=\frac{\overline X_f}{\overline Y}
    -\frac1L\left(
    \frac{\overline X_fs_Y^2}{\overline Y^3}
    -\frac{s_{XY}(f)}{\overline Y^2}
    \right).
\end{aligned}
    \label{eq:main_analytic_ratio_correction}
\end{equation}
\begin{proposition}[Leading bias correction]
\label{prop:main_leading_bias}
Let $M=\min_iM_i$.
\begin{equation}
    \lim_{L,M\to\infty}
    L\!\left(
    \mathbb E[\widetilde T_k(f)]-T_{k,H}(f)
    \right)=0.
    \label{eq:main_leading_bias_proposition}
\end{equation}
This correction mitigates the leading bias introduced by the random denominator when normalizing noisy estimates.
\end{proposition}
The proposition applies to any fixed bounded statistic $f$ of a candidate stroke.
More details about the regularity conditions for Proposition~\ref{prop:main_leading_bias} are provided in Appendix~\ref{app:ratio_bias}.
More details about the finite-candidate importance approximation and the unbiased estimation of unnormalized masses are provided in Appendix~\ref{app:importance}.
More details about the ratio-bias derivation and the unequal fresh-rollout analysis are provided in Appendix~\ref{app:ratio_bias}.
More details about failed rollouts and numerical fallback rules are provided in Appendix~\ref{app:validity}.

In summary, CANVAS converts a joint model-and-render complete-SVG target into a parser-aligned, future-aware stroke navigation rule, and reliably approximates it under a limited generation and rendering budget using finite-horizon rollouts and influence- and cost-aware visual adaptive sampling.

\begin{table*}[t]
\centering
\caption{Quantitative comparison on HeisenVec. The upper block augments the original vHector results~\citep{zini2025vhector} with our consistency evaluations; the lower block reports CANVAS results. Best and second-best results are bold and underlined.}
\label{tab:heisenvec_main}
\vspace{-0.5mm}
\renewcommand{\arraystretch}{0.98}
\setlength{\tabcolsep}{2.6pt}
\resizebox{\textwidth}{!}{%
\begin{tabular}{llrrrrrrrr@{\hspace{5pt}}rrrrrr}
\toprule
\multirow{2}{*}{Backbone} &
\multirow{2}{*}{Type} &
\multirow{2}{*}{\shortstack{Token\\pred.}} &
\multicolumn{6}{c}{Standard metrics} &
\multicolumn{6}{c}{Additional consistency metrics}
\\
\cmidrule(lr){4-9}
\cmidrule(lr){10-15}
& &
& CLIP-S$\uparrow$
& FID$\downarrow$
& HPSv2$\uparrow$
& SSIM$\uparrow$
& LPIPS$\downarrow$
& DINO-S$\uparrow$
& LCI$_{9\times9}\uparrow$
& CVIU-C$\uparrow$
& SC$\uparrow$
& SL$\uparrow$
& CE$\uparrow$
& LaDe$\uparrow$
\\
\midrule
\multicolumn{15}{l}{\textbf{Original vHector Table 2 baseline results}}
\\
\midrule
CodeLLaMA~\citep{zini2025vhector}
& LLM
& 8,192
& 60.82
& 793.97
& 13.83
& 59.51
& 67.79
& 25.68
& 0.437
& 0.271
& 0.314
& 0.122
& 0.290
& 2.134
\\
IconShop~\citep{wu2023iconshop}
& Dec.
& --
& 66.09
& 316.86
& 12.94
& 46.78
& 68.29
& 39.23
& 0.737
& \textbf{0.594}
& 0.688
& 0.283
& 0.633
& 3.691
\\
LLM4SVG~\citep{xing2025llm4svg}
& LLM
& 2,048
& 62.08
& 830.70
& 14.01
& \underline{69.62}
& 61.85
& 35.92
& 0.728
& 0.190
& 0.436
& 0.209
& 0.497
& 2.674
\\
OmniSVG~\citep{yang2025omnisvg}
& VLM
& 8,192
& 63.82
& 309.20
& 14.46
& 61.05
& 65.93
& 37.63
& 0.676
& 0.457
& 0.465
& 0.235
& 0.480
& 2.806
\\
vHector 8B~\citep{zini2025vhector}
& LLM
& 8,192
& \underline{67.84}
& \textbf{105.22}
& \underline{15.32}
& 65.91
& \underline{57.54}
& \underline{49.33}
& 0.723
& 0.480
& 0.531
& \underline{0.360}
& 0.522
& 2.950
\\
vHector 3B~\citep{zini2025vhector}
& LLM
& 8,192
& 64.30
& 343.90
& 14.34
& 63.62
& 61.40
& 37.76
& 0.470
& 0.310
& 0.314
& 0.175
& 0.280
& 2.233
\\
\midrule
\multicolumn{15}{l}{\textbf{Our plug-and-play CANVAS results}}
\\
\midrule
CodeLLaMA-7B+CANVAS
& LLM
& 8,192
& 66.55
& 177.11
& 14.50
& 65.48
& 63.16
& 31.63
& 0.454
& 0.277
& 0.339
& 0.155
& 0.317
& 2.231
\\
IconShop+CANVAS
& Dec.
& --
& 67.70
& 177.99
& 13.59
& 50.97
& 65.46
& 43.84
& \underline{0.746}
& \textbf{0.594}
& \underline{0.725}
& 0.357
& \underline{0.701}
& \underline{3.875}
\\
LLM4SVG+CANVAS
& LLM
& 2,048
& 67.02
& 146.64
& 14.30
& \textbf{71.21}
& 60.59
& 38.51
& 0.635
& 0.172
& 0.445
& 0.253
& 0.503
& 2.732
\\
OmniSVG+CANVAS
& VLM
& 8,192
& 67.24
& \underline{119.76}
& 14.80
& 63.85
& 64.05
& 40.76
& 0.652
& 0.444
& 0.483
& 0.268
& 0.498
& 2.875
\\
vHector-8B+CANVAS
& LLM
& 8,192
& \textbf{72.32}
& 151.63
& \textbf{20.22}
& 59.52
& \textbf{56.07}
& \textbf{59.40}
& \textbf{0.774}
& \underline{0.526}
& \textbf{0.931}
& \textbf{0.927}
& \textbf{0.916}
& \textbf{4.736}
\\
vHector-3B+CANVAS
& LLM
& 8,192
& 65.77
& 150.84
& 15.10
& 65.12
& 59.91
& 40.70
& 0.510
& 0.324
& 0.368
& 0.229
& 0.344
& 2.420
\\
\bottomrule
\end{tabular}%
}
\vspace{-2mm}
\end{table*}

\begin{table*}[t]
\centering
\caption{Quantitative comparison on the IntroSVG evaluation set. The upper block augments the original IntroSVG results~\citep{wang2026introsvg} with our consistency evaluations; the lower block reports CANVAS results. Best and second-best results are bold and underlined.}
\label{tab:introsvg_main}
\vspace{-0.5mm}
\renewcommand{\arraystretch}{0.98}
\setlength{\tabcolsep}{2.6pt}
\resizebox{\textwidth}{!}{%
\begin{tabular}{lrrrrrr@{\hspace{5pt}}rrrrrr}
\toprule
\multirow{2}{*}{Backbone}
& \multicolumn{6}{c}{Standard metrics}
& \multicolumn{6}{c}{Additional consistency metrics}
\\
\cmidrule(lr){2-7}
\cmidrule(lr){8-13}
& \shortstack{Avg.\\Token$\downarrow$}
& RSR$\uparrow$
& FID$\downarrow$
& CLIP-T2I$\uparrow$
& Aesthetic$\uparrow$
& HPS$\uparrow$
& LCI$_{9\times9}\uparrow$
& CVIU-C$\uparrow$
& SC$\uparrow$
& SL$\uparrow$
& CE$\uparrow$
& LaDe$\uparrow$
\\
\midrule
\multicolumn{13}{l}{\textbf{Original IntroSVG baseline results}}
\\
\midrule
OmniSVG (Qwen2.5-3B-Instruct)~\citep{yang2025omnisvg}
& 2260.54
& 75.36
& 142.38
& 0.2297
& 4.7232
& 0.1877
& 0.676
& 0.466
& 0.507
& 0.293
& 0.513
& 2.808
\\
SVGen (Qwen2.5-Coder-7B-Instruct)~\citep{wang2025svgen}
& \underline{1531.42}
& 84.64
& 26.27
& 0.2339
& 4.5858
& 0.1916
& 0.841
& 0.624
& 0.774
& 0.586
& 0.798
& 3.887
\\
IntroSVG~\citep{wang2026introsvg}
& 1707.77
& \underline{99.26}
& \underline{26.18}
& \underline{0.2529}
& \underline{4.8894}
& \underline{0.1969}
& 0.841
& 0.620
& 0.775
& 0.588
& 0.802
& 3.892
\\
\midrule
\multicolumn{13}{l}{\textbf{Our plug-and-play CANVAS result}}
\\
\midrule
OmniSVG+CANVAS
& {1337.52}
& {92.57}
& {25.73}
& {0.2491}
& {4.5970}
& {0.1925}
& 0.839
& 0.625
& 0.773
& 0.582
& 0.804
& 3.858
\\
SVGen+CANVAS
& {1319.20}
& {81.29}
& {25.71}
& {0.2499}
& {4.5917}
& {0.1930}
& \textbf{0.847}
& \textbf{0.634}
& \underline{0.781}
& \underline{0.555}
& \underline{0.799}
& \underline{3.922}
\\
IntroSVG+CANVAS
& \textbf{1416.04}
& \textbf{100.00}
& \textbf{25.66}
& \textbf{0.2615}
& \textbf{4.9872}
& \textbf{0.2008}
& \underline{0.844}
& \underline{0.624}
& \textbf{0.791}
& \textbf{0.613}
& \textbf{0.810}
& \textbf{3.999}
\\
\bottomrule
\end{tabular}%
}
\vspace{-2mm}
\end{table*}

\begin{table*}[t]
\begin{minipage}{\textwidth}
\centering
\captionsetup{type=table}
\caption{Ablation of the decision unit. Times are average generation wall-clock per sample on four H200 GPUs: Ours uses the completed strict four-lane runtime, while the two-sample alternatives are linearly extrapolated. Higher is better except for FID, LPIPS, and time. Best and second-best results are bold and underlined, respectively.}
\label{tab:main_decision_reward_ablation}
\vspace{-0.5mm}
\renewcommand{\arraystretch}{0.98}
\setlength{\tabcolsep}{2.5pt}
\resizebox{\textwidth}{!}{%
\begin{tabular}{lrrrrrr@{\hspace{5pt}}rrrrrr@{\hspace{5pt}}l}
\toprule
\multirow{2}{*}{Variant} &
\multicolumn{6}{c}{Standard metrics} &
\multicolumn{6}{c}{Additional consistency metrics} &
\multirow{2}{*}{\shortstack{Avg. time/sample (s)\\4$\times$H200$\downarrow$}}
\\
\cmidrule(lr){2-7}
\cmidrule(lr){8-13}
&
CLIP-S$\uparrow$
& FID$\downarrow$
& HPSv2$\uparrow$
& SSIM$\uparrow$
& LPIPS$\downarrow$
& DINO-S$\uparrow$
& LCI$_{9\times9}\uparrow$
& CVIU-C$\uparrow$
& SC$\uparrow$
& SL$\uparrow$
& CE$\uparrow$
& LaDe$\uparrow$
&
\\
\midrule
Token-wise
& 68.878
& \textbf{66.574}
& 13.753
& \textbf{69.913}
& 62.766
& 32.981
& \textbf{0.8801}
& 0.3121
& 0.4723
& 0.2663
& 0.5230
& 2.6875
& \(\approx\)7{,}112.4\,s
\\
Fixed block-wise ($m=20$)
& \underline{71.326}
& \underline{89.597}
& \underline{14.395}
& \underline{62.912}
& \underline{61.039}
& \underline{56.761}
& \underline{0.8195}
& \textbf{0.6021}
& \underline{0.6355}
& \underline{0.3769}
& \underline{0.6657}
& \underline{3.7224}
& \underline{\(\approx\)165.9\,s}
\\
\midrule
\textbf{Ours}
& \textbf{72.324}
& 151.634
& \textbf{20.219}
& 59.516
& \textbf{56.073}
& \textbf{59.400}
& 0.7736
& \underline{0.5256}
& \textbf{0.9312}
& \textbf{0.9274}
& \textbf{0.9159}
& \textbf{4.7362}
& \textbf{\(\approx\)17.8\,s}
\\
\bottomrule
\end{tabular}%
}
\vspace{-0.5mm}

\captionsetup{type=table}
\caption{Rollout-horizon ablation on vHector-8B. Times are average generation wall-clock per sample on four H200 GPUs. H1(Ours), H2, and H3 use completed-run timings converted to the same four-card basis (excluding user-requested pauses), while HEOS uses a linear throughput extrapolation. Higher is better except for FID, LPIPS, and time. Best and second-best results are bold and underlined, respectively.}
\label{tab:main_rollout_horizon_ablation}
\vspace{-0.5mm}
\renewcommand{\arraystretch}{0.98}
\setlength{\tabcolsep}{2.5pt}
\resizebox{\textwidth}{!}{%
\begin{tabular}{lrrrrrr@{\hspace{5pt}}rrrrrr@{\hspace{5pt}}l}
\toprule
\multirow{2}{*}{Variant} &
\multicolumn{6}{c}{Standard metrics} &
\multicolumn{6}{c}{Additional consistency metrics} &
\multirow{2}{*}{\shortstack{Avg. time/sample (s)\\4$\times$H200$\downarrow$}}
\\
\cmidrule(lr){2-7}
\cmidrule(lr){8-13}
&
CLIP-S$\uparrow$
& FID$\downarrow$
& HPSv2$\uparrow$
& SSIM$\uparrow$
& LPIPS$\downarrow$
& DINO-S$\uparrow$
& LCI$_{9\times9}\uparrow$
& CVIU-C$\uparrow$
& SC$\uparrow$
& SL$\uparrow$
& CE$\uparrow$
& LaDe$\uparrow$
&
\\
\midrule
H2
& 69.977
& \textbf{102.383}
& 15.806
& \underline{62.604}
& 59.617
& 56.048
& 0.7976
& 0.5811
& 0.5698
& 0.4184
& 0.6177
& 3.1752
& \underline{$\approx$38.2\,s}
\\
H3
& \underline{71.046}
& 198.614
& \underline{17.844}
& \textbf{63.681}
& \textbf{54.196}
& \textbf{65.620}
& \textbf{0.8377}
& \underline{0.6180}
& \underline{0.6951}
& \underline{0.5182}
& \underline{0.7338}
& \underline{3.6844}
& $\approx$63.4\,s
\\
HEOS
& 70.358
& 205.196
& 17.300
& 61.466
& \underline{55.391}
& \underline{59.878}
& \underline{0.8270}
& \textbf{0.6212}
& 0.6456
& 0.4816
& 0.6971
& 3.5956
& $\approx$227.5\,s
\\
\midrule
\textbf{H1(Ours)}
& \textbf{72.324}
& \underline{151.634}
& \textbf{20.219}
& 59.516
& 56.073
& 59.400
& 0.7736
& 0.5256
& \textbf{0.9312}
& \textbf{0.9274}
& \textbf{0.9159}
& \textbf{4.7362}
& \textbf{$\approx$17.8\,s}
\\
\bottomrule
\end{tabular}%
}
\vspace{-2mm}
\end{minipage}
\vspace{-0.5mm}
\end{table*}

\begingroup
\setlength{\parskip}{-1pt}
\section{Experiments}
\label{sec:exp}

\noindent\textbf{Datasets and evaluation metrics.}
To evaluate the effectiveness and generalization ability of CANVAS across different SVG generators, we conduct experiments on two complementary Text-to-SVG benchmarks: the HeisenVec benchmark introduced with vHector~\citep{zini2025vhector} and the benchmark introduced by IntroSVG~\citep{wang2026introsvg}.
Following their official evaluation protocols, we retain each benchmark's original metrics.
HeisenVec targets complex, long-context Text-to-SVG generation and exposes the difficulty of maintaining coherent SVG structure over extended outputs~\citep{zini2025vhector}.
For this dataset, we follow vHector~\citep{zini2025vhector} and report CLIP Score (CLIP-S), Fr\'echet Inception Distance (FID), Human Preference Score v2 (HPSv2), Structural Similarity Index Measure (SSIM), Learned Perceptual Image Patch Similarity (LPIPS), and DINO Similarity (DINO-S).
IntroSVG is a more recent Text-to-SVG framework that explicitly incorporates rendered-image feedback to iteratively improve the visual fidelity and semantic accuracy of complex, multicolor SVGs. Its unified evaluation setting therefore provides a strong complementary testbed for assessing whether CANVAS can further improve global consistency in structurally rich, long-form SVG generation~\citep{wang2026introsvg}.
For this dataset, we follow IntroSVG~\citep{wang2026introsvg} and report average token count (Avg. Token), Render Success Rate (RSR), FID, CLIP text-to-image similarity (CLIP-T2I), Aesthetic Score, and HPS.
Definitions and execution details for the metrics used by both benchmarks are provided in Appendix~\ref{app:benchmark_metrics}.

Beyond these benchmark-standard measures, we introduce six complementary consistency metrics for long-range SVG geometry and composition.
They are the Local Connectivity Index (LCI$_{9\times9}$), CVIU Statistical Complexity (CVIU-C), Structural Coherence (SC), Spatial Logic (SL), Character Efficiency (CE), and LaDe Cross-Layer Consistency (LaDe).
Their formulas, the intuition behind them, SVG transfer protocols, execution details, and qualitative interpretations are provided in Appendix~\ref{app:consistency_metric_visualizations}.

\noindent\textbf{Implementation details.}
Main experiments use four NVIDIA H200 GPUs with $\alpha=2$, $\beta=64$, and $H=1$; further details are provided in Appendices~\ref{app:validity} and~\ref{app:procedure}.

\subsection{Main Results}
\label{sec:main_results}

We apply CANVAS to mainstream Text-to-SVG generators on HeisenVec~\citep{zini2025vhector} and IntroSVG~\citep{wang2026introsvg} (Tabs.~\ref{tab:heisenvec_main} and~\ref{tab:introsvg_main}).
Across both benchmarks and the large majority of reported metrics, CANVAS yields consistent improvements, supporting its effectiveness and generalization ability in long-context SVG generation.
Comparisons with temperature scaling, beam search, and Best-of-$N$, together with qualitative visualizations, are provided in Appendix~\ref{app:ablation_studies}.

\subsection{Ablation Studies}
\label{sec:ablation_studies}

We conduct extensive ablation experiments using vHector-8B and the HeisenVec evaluation protocol from Tab.~\ref{tab:heisenvec_main}.
Additional ablation studies are provided in Appendix~\ref{app:ablation_studies}.
We analyze two choices that are central to making future-aware SVG navigation both effective and computationally practical: the decision unit of CANVAS and the rollout horizon used to estimate the candidates' future values.

\paragraph{Impact of using strokes as the decision unit.}
In CANVAS, we use parser-complete strokes as the decision unit.
To assess the contribution of this choice, we test two variants.
In the first variant (\textbf{token-wise}), CANVAS applies its evaluation and selection procedure at every token.
In the second variant (\textbf{fixed block-wise}), CANVAS applies the same procedure once per fixed-length token block.
As shown in Tab.~\ref{tab:main_decision_reward_ablation}, although these alternatives may lead on individual metrics, their cost--quality tradeoffs are unsatisfactory: token-wise decisions require an unacceptable inference-time cost, while the cost of fixed block-wise decisions can be moved into a practical range by changing the block length $m$, but their performance is unstable because a fixed block may end inside a coordinate, path command, or XML attribute.
We therefore report $m=20$ as a representative setting with comparatively strong performance.
These results support our parser-aligned decision unit: stroke boundaries provide a visually meaningful and manageable action space, striking a favorable balance between computational cost and generation quality.

\paragraph{Impact of visual adaptive sampling.}
For each candidate stroke, CANVAS uses visual adaptive sampling with bounded future rollouts to estimate its future value under a fixed inference-time budget.
To assess the rollout design, we compare the complete one-stroke method (\textbf{H1/Ours}) with two-stroke (\textbf{H2}), three-stroke (\textbf{H3}), and valid-EOS (\textbf{HEOS}) rollouts.
As shown in Tab.~\ref{tab:main_rollout_horizon_ablation}, CANVAS achieves a favorable balance between computational cost and generation quality.
These results demonstrate the effectiveness of our visual adaptive sampling design.

\FloatBarrier
\section{Conclusion}
\label{sec:conclusion}

In this paper, we proposed CANVAS, a novel training-free framework that improves the global consistency of long-context Text-to-SVG decoding by introducing consistency-aware navigation through the SVG generation space.
In CANVAS, we propose several novel designs that enable a fixed autoregressive SVG generator to exploit both its learned preference over complete generation trajectories and direct visual evidence from rendered futures within an acceptable inference-time budget and without any additional training.
Extensive experiments demonstrate the effectiveness and generalization ability of CANVAS.
\endgroup
\FloatBarrier
{\small
\bibliographystyle{ieeenat_fullname}
\bibliography{references}
}

\clearpage
\appendix
\twocolumn[
\begin{center}
  {\LARGE\bfseries Supplementary Material\par}
\end{center}
\vspace{1em}
]

\numberwithin{equation}{section}

\section{How Power Sharpening Concentrates the Complete-Sequence Prior}
\label{app:power_sharpening}

This section defines the complete-sequence power prior used by the target in the main text and proves its trajectory-concentration property. Let $q$ be a prompt, and let $p_\theta$ denote the fixed base model with parameters $\theta$. Let $Y$ be a complete SVG program, let $T(Y)$ denote its number of tokens, and write $Y=(y_1,\ldots,y_{T(Y)})$. For each $t\in\{1,\ldots,T(Y)\}$, $y_t$ is the token at position $t$, and $y_{<t}=(y_1,\ldots,y_{t-1})$ is the prefix preceding that token. Under prompt $q$, the model assigns $Y$ the complete-sequence likelihood $p_\theta(Y\mid q)$ by multiplying the next-token probabilities $p_\theta(y_t\mid y_{<t},q)$, as defined in Eq.~\plaineqref{eq:token_factorization} in the main paper.
Let $\mathcal Y_q$ denote the space of valid, completely terminated SVG programs for prompt $q$; its operational caps are specified in Appendix~\ref{app:boundary_rule}. We define the power-sharpened distribution over this space as
\begin{equation}
\begin{aligned}
    p_{\theta,\alpha}^{\mathrm{pow}}(Y\mid q)
    &=\frac{p_\theta(Y\mid q)^\alpha}{Z_\alpha(q)},
    &\alpha&\geq1,\\
    Z_\alpha(q)
    &=\sum_{Y\in\mathcal Y_q}p_\theta(Y\mid q)^\alpha.
\end{aligned}
    \label{eq:power_prior}
\end{equation}
Here, $\alpha$ controls how strongly the base model's preferences are reinforced at the complete-trajectory level.
Setting $\alpha=1$ leaves the relative probabilities over $\mathcal Y_q$ unchanged, whereas a larger $\alpha$ concentrates the distribution; $Z_\alpha(q)$ is the normalizer that makes the probabilities over $\mathcal Y_q$ sum to one.
Fix prompt $q$ and consider any two complete SVG programs $Y_a$ and $Y_b$ with positive probability; if $p_\theta(Y_a\mid q)>p_\theta(Y_b\mid q)$, then for $\alpha>1$,
\begin{equation}
    \frac{p_{\theta,\alpha}^{\mathrm{pow}}(Y_a\mid q)}
    {p_{\theta,\alpha}^{\mathrm{pow}}(Y_b\mid q)}
    =\left(\frac{p_\theta(Y_a\mid q)}{p_\theta(Y_b\mid q)}\right)^\alpha
    >\frac{p_\theta(Y_a\mid q)}{p_\theta(Y_b\mid q)}.
    \label{eq:power_odds_gap}
\end{equation}
Therefore, power sharpening increases the relative weight of a more likely complete program over a less likely one. The same effect can be studied at the distribution level. Because
\[
p_\theta(Y\mid q)^\alpha
=
\exp\!\left\{\alpha\log p_\theta(Y\mid q)\right\},
\]
the base-model log-likelihood is the natural quantity for describing how the power-sharpened distribution changes with $\alpha$.

Let $\ell(Y)=\log p_\theta(Y\mid q)$ and define
\begin{equation}
\begin{aligned}
m(\alpha)
&=\mathbb E_{Y\sim p_{\theta,\alpha}^{\mathrm{pow}}(\cdot\mid q)}[\ell(Y)]\\
&=\sum_{Y\in\mathcal Y_q}
p_{\theta,\alpha}^{\mathrm{pow}}(Y\mid q)
\log p_\theta(Y\mid q).
\end{aligned}
\end{equation}
Thus, $m(\alpha)$ is the average base-model log-likelihood of a complete SVG program drawn from the power-sharpened distribution. An increase in $m(\alpha)$ means that the sampled programs are assigned higher base-model log-likelihood on average.

Differentiating with respect to $\alpha$ gives
\begin{equation}
\frac{\mathrm d m(\alpha)}{\mathrm d\alpha}
=
\operatorname{Var}_{Y\sim p_{\theta,\alpha}^{\mathrm{pow}}(\cdot\mid q)}
\!\left[\log p_\theta(Y\mid q)\right]
\geq 0.
\label{eq:power_expected_loglikelihood}
\end{equation}
Because a variance is always nonnegative, $m(\alpha)$ cannot decrease as $\alpha$ increases. It increases strictly unless all programs in the support have the same base-model likelihood. Therefore, increasing $\alpha$ shifts probability mass toward complete programs that the base model already considers more likely. This conclusion concerns only model likelihood and does not imply better rendering quality, which motivates the visual reward introduced in the main text.

\section{Unifying Online Decision Units in a Valid SVG Space}
\label{app:boundary_rule}

\paragraph{General block segmentation.}
Here we use a deterministic boundary rule $\Gamma$ to split each generated sequence into blocks, which can describe token-wise, fixed-length, and parser-detected variable-length decision units in a single framework. For any valid complete sequence $Y=(y_1,\ldots,y_{T(Y)})$, the rule produces the unique boundaries
\begin{equation}
\begin{aligned}
    0 &= \tau_0 < \tau_1 < \cdots
       < \tau_{K(Y)} = T(Y),\\
    B_k &= y_{\tau_{k-1}+1:\tau_k},\\
    H_k &= y_{1:\tau_k}.
\end{aligned}
\label{eq:boundary_segmentation}
\end{equation}
Here, $K(Y)$ is the number of blocks in $Y$, $k\in\{1,\ldots,K(Y)\}$ is the block index, and $\tau_k$ is the token position at which the $k$-th block ends, with $\tau_0=0$. Thus, $B_k$ is the $k$-th block, and $H_k=H_{k-1}B_k$ is the full sequence prefix after that block has been appended. The boundary rule operates online: after token $y_t$ is generated, it determines whether the current block ends at position $t$ based on the prefix $y_{1:t}$.

The three decision units are special cases of this rule. For token-wise decisions, each block contains one token, so the $k$-th block ends at token $k$ and $\tau_k=k$. For fixed-length decisions, each block contains $m$ tokens except possibly the final block, so $\tau_k=\min(km,T(Y))$.
For parser-detected decisions, let $h=H_{k-1}$ be the current boundary prefix. Starting from $h$, suppose the incremental SVG parser first detects either a newly completed renderable drawing element or the completion of the entire SVG program after generating a segment $b=(b_1,\ldots,b_{|b|})$. This entire segment, including the token that triggers the event, forms one candidate block. If $b$ is selected and appended to $h$, we set $B_k=b$, $\tau_k=\tau_{k-1}+|b|$, and $H_k=hb$. In this case, no shorter prefix of $b$ already completes a new renderable drawing element or the entire SVG program. That is what the first-hit condition means: appending the complete block $b$ to $h$ triggers $\Gamma$, whereas appending any strict prefix $b_{1:j}$, with $1\leq j<|b|$, does not. 

To support boundary-wise decisions at a general block granularity, including the parser-detected stroke-wise granularity used by CANVAS, we aggregate the base model's token-level probabilities within each candidate block.

Given a current boundary prefix $h$, let $\mathcal B_\Gamma(h)$ denote the candidate blocks whose endpoint is the next boundary selected by $\Gamma$. Here, each $b\in\mathcal B_\Gamma(h)$ satisfies the first-hit condition defined above. For a candidate block $b=(b_1,\ldots,b_{|b|})$, let $b_{<j}=(b_1,\ldots,b_{j-1})$. Juxtaposition denotes sequence concatenation, so $hb_{<j}$ is the complete token prefix before $b_j$ is generated. The raw base-model probability of generating the entire block $b$ from $h$ is
\begin{equation}
    P_\theta^\Gamma(b\mid h,q)
    =\prod_{j=1}^{|b|}p_\theta(b_j\mid hb_{<j},q),
    \qquad b\in\mathcal B_\Gamma(h).
    \label{eq:boundary_block_probability}
\end{equation}
Here, $P_\theta^\Gamma(b\mid h,q)$ is simply the product of their original autoregressive next-token probabilities, with every token conditioned on the full prefix preceding it.

Now consider a valid complete sequence $Y$. The rule $\Gamma$ uniquely partitions $Y$ into consecutive, non-overlapping blocks $B_1,\ldots,B_{K(Y)}$ that together contain every token exactly once. Since $H_{k-1}$ is the complete prefix before block $B_k$, the same complete-sequence likelihood can be written as
\begin{equation}
    p_\theta(Y\mid q)
    =\prod_{t=1}^{T(Y)}p_\theta(y_t\mid y_{<t},q)
    =\prod_{k=1}^{K(Y)}P_\theta^\Gamma(B_k\mid H_{k-1},q).
    \label{eq:boundary_block_factorization}
\end{equation}
The middle expression multiplies the model probabilities one token at a time, whereas the final expression groups exactly the same factors by block. Therefore, changing the boundary rule changes only the decision granularity, not the likelihood assigned to $Y$. When $\Gamma$ is specialized to the parser-detected rule, each drawing block is parser-complete, and the final product gives the stroke-wise factorization used by CANVAS.

\paragraph{From the complete target to the stroke-wise conditional.}
To derive the stroke-wise conditional used by CANVAS, we first write the corresponding conditional for a general boundary rule $\Gamma$. Under the parser-detected rule used by CANVAS, each drawing candidate is a parser-complete stroke. A terminating candidate instead completes the SVG by closing all remaining SVG/XML tags and emitting EOS.

Let $h=H_{k-1}$ be the current committed prefix, and let $b\in\mathcal B_\Gamma(h)$ be a candidate decision unit. Let $\mathcal Y_q$ denote the set of valid, completely terminated SVG programs for prompt $q$ under the fixed operational caps.

For fixed $h$ and $b$, let $C$ range over all future token suffixes that complete $hb$ into an element of $\mathcal Y_q$. Here, $hbC$ denotes the concatenation of $h$, $b$, and $C$. Thus, $C$ contains every token generated after $b$ until valid SVG completion.

For each candidate $b$, we sum the complete-target contributions of all such valid futures and combine this future contribution with the base-model probability of $b$:
\begin{equation}
\begin{aligned}
    G_k^\Gamma(hb,q)
    &=
    \sum_{C:\,hbC\in\mathcal Y_q}
    p_\theta(C\mid hb,q)^\alpha
    e^{\beta s(\mathcal R(hbC),q)},\\
    W_k^\Gamma(b;h,q)
    &=
    P_\theta^\Gamma(b\mid h,q)^\alpha
    G_k^\Gamma(hb,q).
\end{aligned}
\label{eq:appendix_block_future}
\end{equation}
The two quantities in Eq.~\plaineqref{eq:appendix_block_future} separate the contribution of the future continuations from the contribution of the current candidate.

For a fixed candidate $b$, consider one valid future suffix $C$. The factor $p_\theta(C\mid hb,q)^\alpha$ measures how strongly the power-sharpened base model favors this continuation, with $\alpha$ controlling the strength of the sharpening. The complete program $hbC$ is rendered by $\mathcal R$, and $s(\mathcal R(hbC),q)$ is its final visual score. The factor $e^{\beta s(\mathcal R(hbC),q)}$ converts this score into a positive visual reward, with $\beta$ controlling the strength of the visual preference. Their product is the target contribution of this particular complete future. Summing this contribution over all valid suffixes $C$ gives $G_k^\Gamma(hb,q)$, the aggregate future score after candidate $b$.

The factor $P_\theta^\Gamma(b\mid h,q)^\alpha$ measures the power-sharpened base-model preference for generating the current candidate $b$ from prefix $h$. Multiplying it by $G_k^\Gamma(hb,q)$ gives $W_k^\Gamma(b;h,q)$, the unnormalized score assigned to candidate $b$. This score combines how plausible $b$ is under the base model with the likelihoods and visual rewards of all valid complete SVGs that can follow it. It is not yet a probability; it becomes one only after the candidate scores are normalized.

All candidates are compared from the same committed prefix $h$. If the likelihood of this prefix were written explicitly, the full unnormalized score for candidate $b$ would contain
\[
    p_\theta(h\mid q)^\alpha
    W_k^\Gamma(b;h,q).
\]
Because $p_\theta(h\mid q)^\alpha$ is identical for every candidate, it cancels when the candidate scores are normalized. The global normalizer of the complete-SVG target cancels for the same reason. We therefore omit these shared factors.

Normalizing the remaining candidate scores gives the conditional distribution at the current boundary:
\begin{equation}
    \Pi_{\alpha,\beta}^\Gamma(b\mid h,q)
    =\frac{W_k^\Gamma(b;h,q)}
    {\sum_{b'\in\mathcal B_\Gamma(h)}
    W_k^\Gamma(b';h,q)}.
    \label{eq:appendix_block_conditional}
\end{equation}
Thus, a candidate receives high conditional probability when it is both preferred by the base model and leads to high-value complete SVG trajectories.

This derivation holds for any boundary rule $\Gamma$. Under the parser-detected rule used by CANVAS, Eq.~\plaineqref{eq:appendix_block_conditional} becomes the stroke-wise conditional.

\paragraph{A valid stroke follows parser first-hit law.}
During the generation of candidate strokes, a normal stop occurs when the parser first recognizes either a complete drawing element or the completion of the entire SVG program, with all remaining SVG/XML tags closed and EOS emitted. A failure stop occurs when a stroke-length token cap, program-length cap, or block-count cap is reached, or when the parser enters an unrecoverable malformed state. All such failure outcomes are represented by the single absorbing symbol $\bot$.

Let $\overline{\mathcal B}(h)$ contain every explicit first-hit segment, whether or not it is later accepted, together with $\bot$. The stopped generation law is
\begin{equation}
\begin{aligned}
    Q_\theta(b\mid h,q)
    &=
    \prod_{j=1}^{|b|}
    p_\theta(b_j\mid hb_{<j},q),
    &&b\ne\bot,\\
    Q_\theta(\bot\mid h,q)
    &=
    1-
    \sum_{\substack{
        b\in\overline{\mathcal B}(h)\\
        b\ne\bot
    }}
    Q_\theta(b\mid h,q).
\end{aligned}
\label{eq:parser_first_hit_law}
\end{equation}
For an explicit segment $b$, $Q_\theta(b\mid h,q)$ is its original autoregressive generation probability. The residual definition of $Q_\theta(\bot\mid h,q)$ assigns all remaining probability mass to failure, ensuring that $Q_\theta$ sums to one.

Not every stopped outcome should be committed. Let $\mathcal V(h)\subset\overline{\mathcal B}(h)$ contain the parser-valid candidate strokes and the terminating candidate that completes the SVG program. Its total stopped-law mass is
\[
    Q_\theta(\mathcal V(h)\mid h,q)
    =
    \sum_{b\in\mathcal V(h)}
    Q_\theta(b\mid h,q).
\]
Conditioning the stopped law on this accepted set gives the proposal used at decision step $k$:
\begin{equation}
    q_k(b\mid h,q)
    =
    \frac{Q_\theta(b\mid h,q)}
    {Q_\theta(\mathcal V(h)\mid h,q)},
    \qquad
    b\in\mathcal V(h).
    \label{eq:valid_block_proposal}
\end{equation}

This proposal can be sampled by rejecting stopped outcomes outside $\mathcal V(h)$. The denominator is shared by all accepted candidates at the same step and therefore cancels in their normalized importance weights. If $Q_\theta(\mathcal V(h)\mid h,q)=0$, no valid candidate can be generated from $h$, so the decoder reports a parser failure rather than committing an invalid segment.

\paragraph{Finite valid program space.}
To make the complete-sequence target and its normalizer well defined, we restrict generation to a finite set $\mathcal Y_q$. This set contains all parser-valid, fully terminated SVG programs for prompt $q$ that satisfy three limits: at most $T_{\max}$ tokens in total, at most $K_{\max}$ parser-detected segments, and a stroke-length token cap of $B_{\max}$. The last cap means that, after one boundary is committed, the parser may read at most $B_{\max}$ additional tokens while waiting for the next boundary. For a complete program $Y$, $K(Y)$ denotes its number of segments, including the terminating segment that closes the remaining SVG/XML tags and emits EOS. If generation exceeds any limit or enters an unrecoverable malformed state, it terminates in the failure state $\bot$ and does not produce an element of $\mathcal Y_q$. These limits ensure that every rollout stops after finitely many steps and that sums over complete sequences are well defined.

\paragraph{More details about rendering and scoring.}
To make the visual score reproducible, we use the same renderer version, viewport, background, raster resolution, SVG canonicalization rule, and evaluator checkpoint throughout an experiment. For a completed program $Y$, the frozen evaluator's output is mapped by a fixed calibration to $s(\mathcal R(Y),q)\in[0,1]$. Thus, the rendering and score of a given program do not depend on when or during which rollout it is evaluated.

\section{Deriving the Complete-SVG Target from the KL-Regularized Objective}
\label{app:variational_target}

Eq.~\plaineqref{eq:main_variational_objective} in the main paper defines the KL-regularized complete-SVG objective~\citep{markovicvoronov2026sampling}.
We now derive its closed-form optimizer term by term.
To simplify notation, conditioning of all distributions on $q$ is omitted below.
Let $\Pi_{\alpha,\beta}$ denote the following normalized complete-SVG target distribution:
\begin{equation}
\begin{aligned}
    \Pi_{\alpha,\beta}(Y)
    &=\frac{p_{\theta,\alpha}^{\mathrm{pow}}(Y)}
    {C_{\alpha,\beta}(q)}
    \exp\{\beta s(\mathcal R(Y),q)\},\\
    C_{\alpha,\beta}(q)
    &=\sum_{Y\in\mathcal Y_q}
    p_{\theta,\alpha}^{\mathrm{pow}}(Y)\\[-1mm]
    &\qquad\times
    \exp\{\beta s(\mathcal R(Y),q)\}.
\end{aligned}
    \label{eq:gibbs_target}
\end{equation}
Because $\mathcal Y_q$ is finite and the score is bounded, $C_{\alpha,\beta}(q)$ is finite and positive whenever the valid base-model support is nonempty.
For any candidate distribution $\pi$ defined on the same support, write
$\mathcal D_{\alpha,\beta}(\pi)=D_{\mathrm{KL}}(\pi\|\Pi_{\alpha,\beta})$.
Expanding this KL divergence gives
\begin{equation}
\begin{aligned}
    \mathcal D_{\alpha,\beta}(\pi)
    &=D_{\mathrm{KL}}\!\left(
    \pi\middle\|p_{\theta,\alpha}^{\mathrm{pow}}
    \right)\\
    &\quad-\beta\,\mathbb E_{Y\sim\pi}[s(\mathcal R(Y),q)]\\
    &\quad+\log C_{\alpha,\beta}(q).
\end{aligned}
\label{eq:variational_kl_expansion}
\end{equation}
Therefore, the objective in Eq.~\plaineqref{eq:main_variational_objective} of the main paper equals $\log C_{\alpha,\beta}(q)-D_{\mathrm{KL}}(\pi\|\Pi_{\alpha,\beta})$.
The KL divergence is nonnegative and equals zero only when the two distributions are identical, so the unique optimizer is $\pi=\Pi_{\alpha,\beta}$.
Substituting $p_{\theta,\alpha}^{\mathrm{pow}}(Y\mid q)=p_\theta(Y\mid q)^\alpha/Z_\alpha(q)$ and absorbing the $Y$-independent $Z_\alpha(q)$ into the normalizer gives Eq.~\plaineqref{eq:complete_svg_target} in the main paper.

\section{Turning the Complete-SVG Target into an Online Stroke-Wise Navigation Rule}
\label{app:exact_conditional}

To turn the global complete-SVG target into the online navigation rule used by CANVAS, this section derives the conditional distribution of the next stroke given the current committed prefix. The goal is to show that the stroke-wise rule is the conditional induced by the complete-SVG target rather than a separate local heuristic.

The derivation has four steps. We first decompose the complete-render reward into boundary-level visual changes without altering the normalized target. We then separate the contribution of the current candidate from the contributions of all futures that can follow it, rewrite the future contribution as an expectation under the base-model rollout law, and finally normalize the candidate scores. We carry out these steps for a general deterministic online boundary rule $\Gamma$ before specializing it to parser-detected strokes.

\paragraph{Step 1: Decomposing the complete-render reward.}
To assign visual credit at individual boundaries without repeatedly counting the same partial rendering, we express the complete-render reward as a product of incremental visual factors. For the deterministic online boundary rule $\Gamma$ introduced above, let $H_k$ be the sequence prefix after the $k$-th decision unit has been appended, and define
\begin{equation}
    \psi_k(H_k,q)
    =\exp\!\left\{\beta\left[
    s(\mathcal R(H_k),q)-s(\mathcal R(H_{k-1}),q)
    \right]\right\}.
    \label{eq:appendix_render_potential}
\end{equation}
The factor $\psi_k(H_k,q)$ compares the rendered score after the $k$-th decision with the score before that decision. Multiplying these factors over the complete sequence gives
\begin{equation}
    \prod_{k=1}^{K(Y)}\psi_k(H_k,q)
    =\exp\!\left\{\beta\left[
    s(\mathcal R(Y),q)-s(\mathcal R(H_0),q)
    \right]\right\}.
    \label{eq:appendix_telescoping_potential}
\end{equation}
Here, every intermediate score appears once with a positive sign and once with a negative sign, so these scores cancel pairwise. Only the final score $s(\mathcal R(Y),q)$ and the initial-canvas score $s(\mathcal R(H_0),q)$ remain. The initial-canvas score is shared by all programs and therefore cancels when the target is normalized. Thus, the decomposition preserves the complete-render factor in Eq.~\plaineqref{eq:complete_svg_target} and does not give extra reward to a drawing merely because it is split into more decision units. When $\beta>0$, a candidate $b$ appended to prefix $h$ has $\psi_k(hb,q)>1$ if it improves the canvas score, $\psi_k(hb,q)<1$ if it lowers the score, and $\psi_k(hb,q)=1$ if the score is unchanged. When $\beta=0$, all visual factors equal 1 and the target has no visual reweighting.

At this point, the global visual reward has been decomposed into one factor for the current decision and a product of factors for later decisions. We next use this decomposition to group complete trajectories according to their next candidate.

\paragraph{Step 2: Separating the current candidate from its futures.}
To determine the target contribution of a current candidate, we collect the weights of all complete outcomes that begin with that candidate. Let $h=H_{k-1}$ be the current committed prefix, and let $b\in\mathcal B_\Gamma(h)$ be a candidate decision unit beginning at $h$. Appending $b$ produces the new prefix $hb$. For a valid continuation, let $C$ contain every token generated after $b$ until valid SVG termination, so $hbC$ is the corresponding complete program. Any factor that depends only on the shared prefix $h$ is identical for all candidates and is omitted because it cancels during normalization.

We first separate the factors determined by the current candidate from the visual factors determined by its future continuation:
\begin{equation}
\begin{aligned}
    u_k(b)
    &=P_\theta^\Gamma(b\mid h,q)^\alpha\psi_k(hb,q),\\
    \Phi_{>k}(hb,C,q)
    &=\prod_{r=k+1}^{K(hbC)}\psi_r(H_r,q).
\end{aligned}
    \label{eq:current_block_future_potential}
\end{equation}
Here, $u_k(b)$ is the contribution already determined after $b$ has been generated: $P_\theta^\Gamma(b\mid h,q)^\alpha$ is the sharpened base-model preference for $b$, and $\psi_k(hb,q)$ is its immediate visual change. The product $\Phi_{>k}(hb,C,q)$ contains the visual changes of all later decisions along the valid continuation $C$. If $b$ already completes the SVG program, $C$ is empty and there are no later decisions, so this product is empty and equals 1.

At this point, the current and future factors have been separated, but the score of $b$ must still include every future outcome reachable from $hb$. To include the same capped outcomes used by the decoder, let $\overline{\mathcal C}(hb)$ contain all valid terminal suffixes together with the absorbing failure outcome $\bot$. Let $\overline p_\theta(C\mid hb,q)$ be the normalized capped rollout law that continues generation until either valid termination or $\bot$. For a valid suffix, $\Phi_{>k}^{\epsilon}$ is the telescoping product above. For a failure, it is the positive factor defined in Appendix~\ref{app:validity} so that the complete failed trajectory receives reward $\epsilon$. Summing over these outcomes gives
\begin{equation}
\begin{aligned}
    U_k^{\epsilon}(b)
    &=u_k(b)
    \sum_{C\in\overline{\mathcal C}(hb)}
    \overline p_\theta(C\mid hb,q)^\alpha
    \Phi_{>k}^{\epsilon}(hb,C,q)\\
    &=u_k(b)\,V_k^{\epsilon}(hb,q).
\end{aligned}
\label{eq:operational_marginal_mass}
\end{equation}
Here, $U_k^{\epsilon}(b)$ is the unnormalized target mass assigned to all outcomes whose next candidate is $b$. The equality defines $V_k^{\epsilon}(hb,q)$ as the total future contribution after $b$: each outcome is weighted by its power-sharpened suffix likelihood and its future visual factor. The factorization $U_k^{\epsilon}(b)=u_k(b)V_k^{\epsilon}(hb,q)$ therefore separates what is known about the current candidate from what depends on the futures reachable from it.

\paragraph{Step 3: Making future value estimable with rollouts.}
To estimate $V_k^{\epsilon}(hb,q)$ using continuations sampled directly from the capped base model, we rewrite its sum as an expectation under $\overline p_\theta(\cdot\mid hb,q)$. Using
$\overline p_\theta(C\mid hb,q)^\alpha=\overline p_\theta(C\mid hb,q)\,\overline p_\theta(C\mid hb,q)^{\alpha-1}$ gives
\begin{equation}
\begin{aligned}
    V_k^{\epsilon}(hb,q)
    &=\sum_{C\in\overline{\mathcal C}(hb)}
    \overline p_\theta(C\mid hb,q)^\alpha
    \Phi_{>k}^{\epsilon}(hb,C,q)\\
    &=\mathbb E_{C\sim\overline p_\theta(\cdot\mid hb,q)}
    \!\left[
    \overline p_\theta(C\mid hb,q)^{\alpha-1}
    \Phi_{>k}^{\epsilon}(hb,C,q)
    \right].
\end{aligned}
\label{eq:operational_future_value}
\end{equation}
This expectation is an exact rewriting of the preceding sum, not an additional approximation. One factor of $\overline p_\theta$ supplies the sampling distribution, while $\overline p_\theta^{\alpha-1}$ supplies the remaining power needed to preserve the power-sharpened suffix likelihood. Failure outcomes remain in the expectation through their small, strictly positive $\Phi_{>k}^{\epsilon}$ contribution rather than being silently discarded. For $\alpha=1$, $\overline p_\theta^{\alpha-1}$ is defined as 1 on positive-mass suffixes.

At this point, every candidate $b$ has an exact unnormalized score $u_k(b)V_k^{\epsilon}(hb,q)$ that combines its current contribution with the value of all future outcomes. The final step turns these scores into a conditional distribution and chooses the parser rule that defines strokes.

\paragraph{Step 4: Obtaining the stroke-wise selection rule.}
To obtain the probability of selecting the next candidate, we normalize its unnormalized mass over all valid candidates $b\in\mathcal B_\Gamma(h)$:
\begin{equation}
    \Pi_{\alpha,\beta}^{\epsilon}(B_k=b\mid H_{k-1}=h,q)
    =\frac{u_k(b)V_k^{\epsilon}(hb,q)}
    {\sum_{b'\in\mathcal B_\Gamma(h)}u_k(b')V_k^{\epsilon}(hb',q)}.
    \label{eq:operational_boundary_conditional}
\end{equation}
Eq.~\plaineqref{eq:operational_boundary_conditional} holds for any deterministic online boundary rule $\Gamma$ that produces a unique segmentation. The main text omits the $\epsilon$ superscript for concision; as $\epsilon\to0$, the expression recovers the conditional restricted to valid outcomes.

To obtain the stroke-wise rule used by CANVAS, we specialize $\Gamma$ to parser-detected stroke boundaries. For nonterminal decisions in the default path-normalized representation, the next boundary is triggered when the parser recognizes a complete path-wise drawing element; if the representation retains other visible primitives, each complete visible drawing element triggers a boundary in the same way. The terminating boundary occurs when the remaining SVG/XML tags are closed and EOS is emitted, as defined above. This specialization changes only where online decisions are made. It does not change the complete-SVG target or the marginalization that produced the conditional. Together with the definitions of $u_k$ and $V_k^{\epsilon}$ above, Eq.~\plaineqref{eq:operational_boundary_conditional} yields Eqs.~\plaineqref{eq:stroke_factors} and~\plaineqref{eq:stroke_conditional} in the main paper.

\section{Estimating Future Value with Finite-Horizon Rollouts}
\label{app:finite_horizon}

To estimate a candidate's future value with bounded computation, we stop each continuation after at most $H$ future strokes and average several rollouts. We first define this finite-horizon target, then distinguish truncation error from finite-sample error. In the main text, we use simplified notation. Here, \(\overline p_\theta\) denotes the rollout distribution after applying the stopping caps, and the superscript \(\epsilon\) indicates that failed rollouts receive the small positive reward \(\epsilon\).

\paragraph{Bounding each rollout with a finite horizon.}
To bound the cost of one rollout, consider decision step $k$, let $h=H_{k-1}$ be the current committed prefix, and fix a sampled candidate stroke $B_i$. Starting from the new prefix $hB_i$, generate until $H$ additional strokes have been produced or an earlier SVG-termination or failure event occurs. Let $C_{1:H}$ denote the resulting future segment. If the SVG terminates or the rollout fails before reaching \(H\) future strokes, \(C_{1:H}\) contains only the strokes generated before that stopping event.

Under the capped rollout process, \(\overline p_\theta(C_{1:H}\mid hB_i,q)\) is the probability that the base model generates the stopped continuation \(C_{1:H}\) from \(hB_i\). We associate this continuation with a visual factor \(\Phi_{i,H}^{\epsilon}\). For a valid continuation, this factor is the product of its incremental visual factors \(\psi\). If the rollout fails, we instead use the positive factor corresponding to the small failure reward \(\epsilon\) defined in Appendix~\ref{app:validity}. We define the contribution of one rollout and its population finite-horizon value by
\begin{equation}
\begin{aligned}
    Z_{i,H}
    &=\overline p_\theta(C_{1:H}\mid hB_i,q)^{\alpha-1}
    \Phi_{i,H}^{\epsilon},\\
    V_{k,H}^{\epsilon}(hB_i,q)
    &=\mathbb E[Z_{i,H}\mid B_i].
\end{aligned}
    \label{eq:operational_finite_horizon_value}
\end{equation}
Here, $\alpha\geq1$ is the likelihood-sharpening exponent. After one factor of $\overline p_\theta$ is used to sample the rollout, $\overline p_\theta^{\alpha-1}$ supplies the remaining likelihood power and $\Phi_{i,H}^{\epsilon}$ supplies the visual contribution. Thus, $Z_{i,H}$ is one rollout's contribution, $V_{k,H}^{\epsilon}$ is its expectation, and $Z_{i,H}^{(m)}$ denotes the $m$-th independent copy. If all continuations stop within $H$ strokes, $V_{k,H}^{\epsilon}$ equals the exact value $V_k^{\epsilon}$; otherwise, it is a truncated surrogate.

\paragraph{Quantifying finite-horizon truncation error.}
To measure only the approximation introduced by stopping at $H$, consider a rollout that has not yet terminated at the horizon. Let $C_{>H}$ denote all strokes generated after $C_{1:H}$, and let $\Phi_{>k+H}^{\epsilon}$ be their future visual factor. After fixing the continuation \(C_{1:H}\) observed up to the horizon, we average over all possible suffixes that could follow it. We denote their expected additional multiplicative contribution by \(W_H(C_{1:H})\).
\begin{equation}
    W_H(C_{1:H})
    =\mathbb E\!\left[
    \overline p_\theta(C_{>H}\mid hB_iC_{1:H},q)^{\alpha-1}
    \Phi_{>k+H}^{\epsilon}
    \,\middle|\,C_{1:H}\right].
    \label{eq:tail_multiplier}
\end{equation}
The expectation in $W_H$ is taken over the capped base-model law of the tail $C_{>H}$ conditional on $C_{1:H}$. The likelihood term supplies the remaining power for this omitted suffix, while $\Phi_{>k+H}^{\epsilon}$ supplies its visual contribution. If the rollout terminates within the horizon, there is no omitted tail and we set $W_H=1$. The exact future value can then be recovered by multiplying the observed contribution $Z_{i,H}$ by this tail multiplier and taking an expectation. By the tower property,
\begin{equation}
    V_k^{\epsilon}=\mathbb E[Z_{i,H}W_H],
    \qquad
    |V_k^{\epsilon}-V_{k,H}^{\epsilon}|
    \leq\mathbb E\!\left[Z_{i,H}|W_H-1|\right].
    \label{eq:horizon_error_bound}
\end{equation}
where $V_k^{\epsilon}$ is the exact untruncated value for the same candidate. The bound shows directly that truncation error is small when the omitted multiplier $W_H$ stays close to 1 on rollouts with substantial contribution. Because truncating the rollout may change the candidate value, we empirically evaluate different rollout horizons in Appendix~\ref{app:horizon_ablation}.

At this point, the effect of truncating a rollout has been isolated through $W_H$. We next estimate the resulting finite-horizon value without introducing additional selection bias.

\paragraph{Keeping the finite-horizon estimate unbiased.}
To estimate $V_{k,H}^{\epsilon}(hB_i,q)$, we choose the rollout count $M_i$ before observing the fresh rollout outcomes and average $M_i$ independent contributions:
\begin{equation}
    \widehat V_{i,H}
    =\frac1{M_i}\sum_{m=1}^{M_i}Z_{i,H}^{(m)},
    \qquad
    \mathbb E[\widehat V_{i,H}\mid B_i]
    =V_{k,H}^{\epsilon}(hB_i,q).
    \label{eq:operational_rollout_unbiasedness}
\end{equation}
Here, $V_{k,H}^{\epsilon}(hB_i,q)$ is the exact finite-horizon future value of candidate $B_i$ under the capped rollout law. Its sample estimate $\widehat V_{i,H}$ is the average of $M_i$ independent fresh-rollout contributions. Conditional on $B_i$ and the preselected $M_i$, this estimate is unbiased. Thus, $H$ controls truncation of the target, whereas $M_i$ controls Monte Carlo variance. As described in Appendix~\ref{app:allocation}, pilot rollouts are used to determine $M_i$. Once $M_i$ is fixed, independent fresh rollouts are generated to compute $\widehat V_{i,H}$.

\section{Estimating the Next-Stroke Distribution with Importance Sampling}
\label{app:importance}

To avoid enumerating every parser-valid next stroke, we sample finitely many candidates and correct their proposal probabilities with importance weights. We then form unbiased estimates of the unnormalized candidate masses and identify the separate error introduced by normalization. To keep the notation concise, we omit the superscript \(\epsilon\) in this section. Failed rollouts still receive the small positive reward \(\epsilon\) defined in Appendix~\ref{app:validity}.

\paragraph{Recovering target weights with importance sampling.}
To approximate the target without enumerating every parser-valid next stroke, we draw candidates from a proposal distribution $q_k(\cdot\mid h,q)$. Here, $q_k(b\mid h,q)$ is the actual probability that the decoder proposes stroke $b$ from the current prefix $h$. Because candidates with larger proposal probabilities appear more often, we use importance sampling~\citep{kahn1953methods} to correct for these unequal sampling probabilities when computing their target weights.

For this correction to be valid, every candidate with positive target weight must also have positive probability under $q_k$. We then draw $L$ candidates independently, $B_i\overset{\mathrm{iid}}{\sim}q_k(\cdot\mid h,q)$, and define
\begin{equation}
    A_i
    =\frac{u_k(B_i)}{q_k(B_i\mid h,q)},
    \qquad i=1,\ldots,L.
    \label{eq:appendix_candidate_importance}
\end{equation}
We refer to each draw $B_i$ as a candidate particle. The numerator $u_k(B_i)=P_\theta^{\mathrm{stroke}}(B_i\mid h,q)^\alpha\psi_k(hB_i,q)$ is the current-stroke factor required by the target, while the denominator $q_k(B_i\mid h,q)$ is the probability with which that particle was sampled. Thus, $A_i$ corrects the proposal frequency so that proposal-weighted averages recover sums under the unnormalized stroke target.

\paragraph{From corrected weights to the next-stroke distribution.}
To convert the corrected weights into probabilities for choosing the next stroke, we normalize their target contributions over all candidates. We write this normalization using a candidate-level function $f$ whose values are bounded, so that the same formulation covers both individual candidate probabilities and other weighted averages. For a generic candidate $B$, let $A(B)=u_k(B)/q_k(B\mid h,q)$ denote the importance factor defined above. We then define
\begin{equation}
\begin{aligned}
    N_{k,H}(f)
    &=\mathbb E_{B\sim q_k}\!\Bigl[\\[-1mm]
    &\qquad A(B)V_{k,H}(hB,q)f(B)\Bigr],\\
    D_{k,H}&=N_{k,H}(1),\\
    T_{k,H}(f)&=\frac{N_{k,H}(f)}{D_{k,H}}.
\end{aligned}
    \label{eq:population_ratio_target}
\end{equation}
Here, $N_{k,H}(f)$ is the total candidate weight after multiplying each candidate's contribution by $f(B)$, while $D_{k,H}=N_{k,H}(1)$ is the total weight of all candidates. Their ratio $T_{k,H}(f)=N_{k,H}(f)/D_{k,H}$ is therefore the corresponding normalized quantity. In particular, if $f(B)=\mathbf1[B=b]$, only candidate $b$ contributes to the numerator, so $T_{k,H}(f)$ is the probability assigned to $b$ by the finite-horizon next-stroke distribution.

To estimate these quantities from the $L$ sampled particles, we replace each exact future value by its independent fresh-rollout estimate $\widehat V_{i,H}$:
\begin{equation}
    \widehat N_{k,H}(f)
    =\frac1L\sum_{i=1}^{L}A_i\widehat V_{i,H}f(B_i),
    \qquad
    \widehat D_{k,H}=\widehat N_{k,H}(1).
    \label{eq:unnormalized_estimators}
\end{equation}
Here, $N_{k,H}(f)$ and $D_{k,H}$ are the exact numerator and denominator of the finite-horizon normalized statistic. Their finite-sample estimates, $\widehat N_{k,H}(f)$ and $\widehat D_{k,H}$, are obtained by averaging over the $L$ sampled candidates and replacing each exact future value with its independent fresh-rollout estimate $\widehat V_{i,H}$. The proposal correction in $A_i$ removes candidate-sampling bias, while Eq.~\plaineqref{eq:operational_rollout_unbiasedness} removes bias from estimating each finite-horizon future value. Applying iterated expectation therefore gives
\begin{equation}
    \mathbb E[\widehat N_{k,H}(f)]=N_{k,H}(f),
    \qquad
    \mathbb E[\widehat D_{k,H}]=D_{k,H}.
    \label{eq:unnormalized_unbiasedness}
\end{equation}
Thus, candidate sampling and fresh rollouts are unbiased before normalization. Their ratio can nevertheless be biased because $\widehat D_{k,H}$ is random; Appendix~\ref{app:ratio_bias} quantifies and corrects its leading bias.

At this point, the finite particle estimator is well defined for a proposal with full target support. We next state two implementation requirements needed to preserve that interpretation.

\paragraph{Ensuring valid importance weights.}
To make the importance-sampling correction match the actual sampling process, we keep every draw as a separate sample, even when multiple draws produce the same token sequence. The denominator in Eq.~\plaineqref{eq:appendix_candidate_importance} must also use the candidate's actual sampling probability after top-$p$, top-$K$, parser screening, or rejection sampling. If any candidate with positive target weight is assigned zero sampling probability, the estimator no longer represents the full candidate distribution and instead targets only the retained candidates.

Under the default parser-valid proposal in Eq.~\plaineqref{eq:valid_block_proposal}, every candidate with positive target weight also has positive proposal probability, so the full-support condition holds. Optional top-$K$ or other hard screening instead changes the target to the restricted conditional described next.

\paragraph{Separating support restriction from Monte Carlo error.}
To separate the effect of intentionally excluding candidates from the error caused by drawing only finitely many samples, let $\mathcal G_k(h)$ denote the retained candidate set, such as a top-$K$ set. Normalizing Eq.~\plaineqref{eq:operational_boundary_conditional} over the complete candidate space gives the exact one-step conditional, whereas normalizing only over $\mathcal G_k(h)$ gives that conditional further conditioned on $B_k\in\mathcal G_k(h)$. The difference between these two targets is support-restriction error. It is distinct from finite-$L$ candidate Monte Carlo error, finite-$H$ horizon error, and ratio bias.

This distinction concerns one stroke-wise decision; it does not by itself provide an error bound for the complete generated SVG.

\section{Spending the Rollout Budget Where It Most Improves the Decision}
\label{app:allocation}

To use a limited rollout budget effectively, CANVAS gives more samples to candidates that are uncertain, influential in the normalized decision, and inexpensive to evaluate. We estimate these quantities with pilot rollouts, derive the corresponding cost-constrained allocation, and then implement it with the pilot estimates. As in Appendix~\ref{app:importance}, we omit the superscript \(\epsilon\) in this section; all future-value and decision quantities remain defined using the same small positive failure reward.

\paragraph{Using pilot rollouts to plan the fresh-rollout budget.}
To learn how difficult each candidate is before assigning the main budget, we first identify the population quantities that the pilot stage must estimate. For candidate $B_i$, let $V_i=V_{k,H}(hB_i,q)$ be its exact finite-horizon future value, let $\sigma_i^2=\operatorname{Var}(Z_{i,H}^{(m)}\mid B_i)$ be the variance of one rollout contribution, and let $c_i=\mathbb E[c_i^{(m)}\mid B_i]$ be the expected cost of one rollout. Here, $Z_{i,H}^{(m)}$ and $c_i^{(m)}$ denote the finite-horizon contribution and token-generation-plus-rendering cost of any single rollout $m$. We estimate these quantities by drawing the same small number $m_0\geq2$ of pilot futures for every candidate and computing
\begin{equation}
\begin{aligned}
    \widehat V_i^{(0)}
    &=\frac1{m_0}\sum_{m=1}^{m_0}Z_{i,H}^{(m)},\\
    \widehat\sigma_i^2
    &=\frac1{m_0-1}\sum_{m=1}^{m_0}
    \bigl(Z_{i,H}^{(m)}-\widehat V_i^{(0)}\bigr)^2,\\
    \widehat c_i
    &=\frac1{m_0}\sum_{m=1}^{m_0}c_i^{(m)}.
\end{aligned}
\label{eq:pilot_diagnosis}
\end{equation}
The pilot statistics $\widehat V_i^{(0)}$, $\widehat\sigma_i^2$, and $\widehat c_i$ estimate $V_i$, $\sigma_i^2$, and $c_i$, respectively. The parenthesized superscript $(0)$ labels quantities computed during the preliminary pilot stage; it is a stage index rather than an exponent. In particular, $\widehat V_i^{(0)}$ is used only to allocate fresh rollouts, whereas $\widehat V_{i,H}$ is the final estimate computed from those fresh rollouts.

The pilot rollouts are used only to choose the number $M_i$ of fresh rollouts assigned to each candidate; their outcomes are not included in the final estimate $\widehat V_{i,H}$. After $M_i$ is fixed, independent fresh rollouts are generated to compute the final future-value estimate. This separation prevents the same random pilot fluctuation from both changing the sample count and influencing the final estimate.

\paragraph{Identifying which candidates benefit most from additional rollouts.}
To decide where an additional rollout is most useful, we measure how an error in each future value changes the normalized candidate probabilities. Condition on the sampled candidates, and let $D=\sum_iA_iV_i$ be their total unnormalized mass and $\rho_j=A_jV_j/D$ be the ideal normalized probability of candidate $j$. Because $\widehat V_{i,H}$ averages $M_i$ independent fresh rollouts, $\operatorname{Var}(\widehat V_{i,H}\mid B_i)=\sigma_i^2/M_i$.

The sensitivity of probability $\rho_j$ to a perturbation in $V_i$ is
\begin{equation}
    \frac{\partial\rho_j}{\partial V_i}
    =\frac{A_i}{D}\left(\mathbf 1[i=j]-\rho_j\right).
    \label{eq:allocation_sensitivity}
\end{equation}
where $\mathbf1[i=j]$ equals 1 when $i=j$ and 0 otherwise. The factor $A_i/D$ measures the scale of candidate $i$ in the normalized decision, while $\mathbf1[i=j]-\rho_j$ records how changing $V_i$ redistributes probability across coordinate $j$.

To summarize the effect over the full probability vector, apply first-order error propagation and sum over all coordinates:
\begin{equation}
    \mathbb E\|\widehat{\boldsymbol\rho}-\boldsymbol\rho\|_2^2
    \simeq\sum_{i=1}^{L}\frac{d_i}{M_i},
    \qquad
    d_i=\frac{A_i^2\sigma_i^2}{D^2}
    \|\mathbf e_i-\boldsymbol\rho\|_2^2.
    \label{eq:appendix_variance_surrogate}
\end{equation}
Here, $\boldsymbol\rho$ is the ideal candidate-probability vector obtained from the exact finite-horizon values $V_i$, with $\rho_i=A_iV_i/D$. The vector $\widehat{\boldsymbol\rho}$ estimates $\boldsymbol\rho$ by replacing each $V_i$ with its fresh-rollout estimate $\widehat V_{i,H}$. The expectation is taken over the fresh-rollout randomness conditional on the sampled candidates, and $\mathbf e_i$ is the one-hot vector for candidate $i$. The coefficient $d_i$ combines rollout variance $\sigma_i^2$, importance scale $A_i^2/D^2$, and the candidate's influence $\|\mathbf e_i-\boldsymbol\rho\|_2^2$. Candidate $i$ therefore contributes approximately $d_i/M_i$ to the squared decision error, so variance alone is not sufficient for allocation.

At this point, the statistical value of assigning rollouts to each candidate has been summarized by $d_i$. We next balance that value against the cost of generating and rendering those rollouts.

\paragraph{Minimizing decision error under a fixed rollout budget.}
To minimize the decision-error surrogate under a fixed expected budget, let $c_i>0$ be the expected token-generation-plus-rendering cost of one fresh rollout for candidate $i$, and let $C_k$ be the fresh-rollout budget for the current stroke-wise decision. If the population values $d_i$ and $c_i$ were known, the continuous allocation would solve
\begin{equation}
    \min_{M_i>0}\sum_i\frac{d_i}{M_i}
    \quad\text{s.t.}\quad
    \sum_i c_iM_i=C_k.
    \label{eq:oracle_allocation_problem}
\end{equation}
The objective is the approximate squared decision error, and the constraint charges each of the $M_i$ rollouts its expected cost $c_i$. Introducing a multiplier $\lambda$ gives the stationarity condition $-d_i/M_i^2+\lambda c_i=0$. Solving this condition and enforcing the budget yields
\begin{equation}
    M_i^*
    =\frac{C_k\sqrt{d_i/c_i}}
    {\sum_j\sqrt{d_jc_j}},
    \qquad
    \min_{\{M_i\}}\sum_i\frac{d_i}{M_i}
    =\frac{\left(\sum_i\sqrt{c_id_i}\right)^2}{C_k}.
    \label{eq:oracle_allocation_solution}
\end{equation}
The optimal count satisfies $M_i^*\propto\sqrt{d_i/c_i}$: a candidate receives more rollouts when its decision-error coefficient is larger and fewer when each rollout is more expensive. The second expression is the minimum attainable value of the first-order error surrogate under budget $C_k$.

\paragraph{Comparing adaptive and uniform allocation.}
To compare with equal allocation at the same expected cost, set the shared count to $M=C_k/\sum_i c_i$. Its surrogate error is $\mathcal E_{\mathrm{uniform}}=(\sum_i d_i)(\sum_i c_i)/C_k$, whereas Eq.~\plaineqref{eq:oracle_allocation_solution} gives $\mathcal E_{\mathrm{adaptive}}=(\sum_i\sqrt{c_id_i})^2/C_k$. By Cauchy--Schwarz, $\mathcal E_{\mathrm{adaptive}}\leq\mathcal E_{\mathrm{uniform}}$, with equality only when every candidate has the same ratio $d_i/c_i$. This proves Eq.~\plaineqref{eq:main_uniform_adaptive_comparison} in the main paper under the stated first-order surrogate.

\paragraph{Turning the oracle allocation into a practical rule.}
To use the oracle rule during decoding, replace the unknown $V_i$, $\sigma_i^2$, and $c_i$ by the pilot estimates from Eq.~\plaineqref{eq:pilot_diagnosis}. First compute
\begin{equation}
\begin{aligned}
    D^{(0)}&=\sum_iA_i\widehat V_i^{(0)},\\
    \rho_i^{(0)}&=\frac{A_i\widehat V_i^{(0)}}{D^{(0)}},\\
    \widehat d_i
    &=\frac{A_i^2\widehat\sigma_i^2}{(D^{(0)})^2}\\[-1mm]
    &\qquad\times
    \|\mathbf e_i-\boldsymbol\rho^{(0)}\|_2^2.
\end{aligned}
    \label{eq:pilot_plugin_coefficients}
\end{equation}
Here, $d_i$ is the ideal decision-error coefficient defined in Eq.~\plaineqref{eq:appendix_variance_surrogate}. Replacing the exact future values $V_i$ with their pilot estimates $\widehat V_i^{(0)}$ gives the pilot total mass $D^{(0)}$ and the pilot probability vector $\boldsymbol\rho^{(0)}$, whose $i$-th coordinate is $\rho_i^{(0)}$. The parenthesized superscript $(0)$ indicates that $D^{(0)}$ and $\boldsymbol\rho^{(0)}$ are computed from pilot-stage estimates; it is a stage label rather than an exponent. Combining these pilot quantities with $\widehat\sigma_i^2$ gives $\widehat d_i$, the pilot plug-in estimate of $d_i$. These quantities determine only the allocation; they do not enter the final fresh-rollout future-value estimate.

To prevent any candidate from receiving too few or too many rollouts, we impose $M_{\min}\leq M_i\leq M_{\max}$. The continuous Karush--Kuhn--Tucker solution is
\begin{equation}
    M_i(\lambda)
    =\operatorname{clip}\!\left(
    \sqrt{\frac{\widehat d_i}{\lambda\widehat c_i}},
    M_{\min},M_{\max}
    \right),
    \label{eq:clipped_kkt_allocation}
\end{equation}
Here, \(\operatorname{clip}(x,M_{\min},M_{\max})\) keeps \(x\) unchanged when it lies within the allowed interval, sets it to \(M_{\min}\) when it is too small, and sets it to \(M_{\max}\) when it is too large. The scalar \(\lambda\) controls the total allocation: increasing \(\lambda\) decreases the rollout counts. We choose \(\lambda\) by a one-dimensional monotone search so that the estimated total cost \(\sum_i \widehat c_i M_i(\lambda)\) stays within the budget \(C_k\) and uses as much of that budget as the bounds allow.

Because rollout counts must be integers, we round the continuous allocation by repeatedly assigning an additional rollout where it gives the largest surrogate reduction per unit cost. If candidate $i$ currently has $M_i$ rollouts, that marginal gain is
\begin{equation}
    \Delta_i(M_i)
    =\frac{\widehat d_i}{\widehat c_i}
    \left(\frac1{M_i}-\frac1{M_i+1}\right).
    \label{eq:rounding_marginal_gain}
\end{equation}
To obtain integer rollout counts, we start from \(M_i=M_{\min}\) for every candidate and add rollouts one at a time. If candidate \(i\) currently has \(M_i\) rollouts, then
\[
\widehat d_i\left(\frac{1}{M_i}-\frac{1}{M_i+1}\right)
\]is the estimated reduction in the decision-error surrogate produced by one additional rollout. Dividing this reduction by the estimated rollout cost \(\widehat c_i\) gives \(\Delta_i(M_i)\), the estimated benefit per unit cost. At each iteration, we assign the next affordable rollout to the candidate with the largest \(\Delta_i(M_i)\). We stop when the remaining budget cannot fund another rollout or every candidate has reached \(M_{\max}\). Ties are broken randomly without using any fresh-rollout outcomes.
To keep the allocation well defined in numerical edge cases, we first compute the minimum required cost \(C_{\min}=\sum_i\widehat c_iM_{\min}\). If \(C_{\min}\) exceeds the available budget, no allocation can satisfy the lower bound, so the fresh-rollout stage is skipped. Otherwise, every candidate receives at least \(M_{\min}\) rollouts, and any zero cost estimate is replaced by a small positive value. If the pilot denominator \(D^{(0)}\) is zero or all \(\widehat d_i\) values are zero, the pilot results provide no usable basis for prioritizing candidates, so we use a uniform allocation instead. The budget is defined in terms of estimated generation-and-rendering cost rather than elapsed wall-clock time, while the rollout caps bound the amount of work in each rollout. Finally, \(L\) is fixed before candidate sampling, and all \(M_i\) values are fixed from the pilot data before fresh rollouts begin. Fresh-rollout outcomes therefore cannot change their own sample counts, preserving the conditional unbiasedness of the final sample means.

\section{Correcting the Leading Bias from Random Normalization}
\label{app:ratio_bias}

To turn the unnormalized estimates from Appendix~\ref{app:importance} into probabilities, we divide by an estimated total mass. Because a random ratio generally satisfies $\mathbb E[\widehat N/\widehat D]\ne N/D$, this normalization introduces bias. We identify its leading $O(L^{-1})$ term and estimate that term from the sampled particles. The resulting correction removes the leading asymptotic bias, but is not exactly unbiased at finite sample sizes. As in Appendix~\ref{app:importance}, we omit the superscript \(\epsilon\), although all quantities remain defined under the same failure-handling rule.

\paragraph{The source of normalization bias explicit.}
To identify where normalization bias comes from, we write each particle's estimated numerator and denominator contributions explicitly. For a bounded candidate statistic $f$, define
\begin{equation}
\begin{aligned}
    Y_i&=A_i\widehat V_{i,H},\\
    X_i(f)&=f(B_i)Y_i,\\
    \overline Y&=\frac1L\sum_{i=1}^{L}Y_i,\\
    \overline X_f&=\frac1L\sum_{i=1}^{L}X_i(f).
\end{aligned}
    \label{eq:ratio_particle_moments}
\end{equation}
Here, $A_i$ is the candidate importance factor and $\widehat V_{i,H}$ is its fresh-rollout future-value estimate. Their product $Y_i$ is the estimated unnormalized mass of particle $B_i$. Multiplying by $f(B_i)$ gives its numerator contribution $X_i(f)$. The averages $\overline Y$ and $\overline X_f$ are therefore the estimated denominator and numerator of the normalized statistic.

To quantify the second-order fluctuations that cause ratio bias, assume $L\geq2$ and compute the sample variance of the denominator contributions and their sample covariance with the numerator contributions:
\begin{equation}
\begin{aligned}
    s_Y^2&=\frac1{L-1}\sum_i(Y_i-\overline Y)^2,\\
    s_{XY}(f)&=\frac1{L-1}\sum_i
    (X_i(f)-\overline X_f)(Y_i-\overline Y).
\end{aligned}
    \label{eq:ratio_sample_moments}
\end{equation}
The quantity $s_Y^2$ measures particle-to-particle variation in the estimated denominator mass, while $s_{XY}(f)$ measures how numerator and denominator fluctuations move together.

\paragraph{Identifying the leading normalization bias.}
To define the population bias before estimating it, let $\mu_X=\mathbb E[\overline X_f]$ and $\mu_Y=\mathbb E[\overline Y]$. Also define the scaled denominator variance $\nu_{Y,L}=L\operatorname{Var}(\overline Y)$ and scaled numerator--denominator covariance $\nu_{XY,L}=L\operatorname{Cov}(\overline X_f,\overline Y)$. A second-order delta expansion of the random ratio $\overline X_f/\overline Y$ gives
\begin{equation}
\begin{aligned}
    B_L(f)&=\frac1L\left(
    \frac{\mu_X\nu_{Y,L}}{\mu_Y^3}
    -\frac{\nu_{XY,L}}{\mu_Y^2}
    \right),\\
    \mathbb E\!\left[\frac{\overline X_f}{\overline Y}\right]-T_{k,H}(f)
    &=B_L(f)+o(L^{-1}).
\end{aligned}
    \label{eq:leading_ratio_bias}
\end{equation}
Candidate symmetry, conditionally unbiased fresh rollouts, and the pilot/fresh split imply $\mu_X/\mu_Y=T_{k,H}(f)$. Thus, $T_{k,H}(f)$ is the ideal finite-horizon normalized target, while $B_L(f)$ is the leading $O(L^{-1})$ difference between that target and the expected random ratio. The notation $o(L^{-1})$ denotes a smaller-order remainder.

\paragraph{Estimating and removing the leading bias.}
To estimate the normalized target and its leading bias from the observed particles, write
\begin{equation}
\begin{aligned}
    \widehat T_k^+(f)&=\frac{\overline X_f}{\overline Y},\\
    \widehat B_L(f)&=\frac1L\left(
    \frac{\overline X_fs_Y^2}{\overline Y^3}
    -\frac{s_{XY}(f)}{\overline Y^2}
    \right).
\end{aligned}
    \label{eq:ratio_correction_components}
\end{equation}
Here, $\widehat T_k^+(f)$ is the ordinary self-normalized estimate of $T_{k,H}(f)$. The quantity $\widehat B_L(f)$ estimates the population bias term $B_L(f)$ by combining the observed denominator variance and numerator--denominator covariance, scaled by $1/L$.

The observed moments $s_Y^2$ and $s_{XY}(f)$ consistently estimate the corresponding population second-order quantities. Consequently,
\begin{equation}
    \widehat B_L(f)-B_L(f)=o_p(L^{-1}).
    \label{eq:correction_consistency}
\end{equation}
Eq.~\plaineqref{eq:correction_consistency} means that, in probability, the error in the estimated correction is asymptotically smaller than the $L^{-1}$ bias term being removed.
CANVAS therefore uses $\widetilde T_k(f)=\widehat T_k^+(f)-\widehat B_L(f)$, which is Eq.~\plaineqref{eq:main_analytic_ratio_correction} in the main paper.

At this point, the leading ratio bias caused by finite candidate sampling and its observable correction have been identified. It remains to verify that using different fresh-rollout counts across candidates does not invalidate the expansion.

\paragraph{Preserving the correction under unequal rollout counts.}
To verify that adaptive, unequal rollout counts are covered by the correction, let $M=\min_iM_i$. If one rollout for candidate $i$ has variance $\sigma_i^2$, then its sample-mean future estimate contributes variance $\sigma_i^2/M_i$. The resulting fresh-rollout components of the scaled moments are
\begin{equation}
\begin{aligned}
    \nu_{Y,L}^{\mathrm{future}}
    &=\frac1L\sum_{i=1}^{L}
    \mathbb E\!\left[\frac{A_i^2\sigma_i^2}{M_i}\right]
    =O(M^{-1}),\\
    \nu_{XY,L}^{\mathrm{future}}
    &=\frac1L\sum_{i=1}^{L}
    \mathbb E\!\left[\frac{f(B_i)A_i^2\sigma_i^2}{M_i}\right]
    =O(M^{-1}).
\end{aligned}
    \label{eq:unequal_rollout_moments}
\end{equation}
Here, the expectations include the candidate and pilot randomness determining $A_i$, $\sigma_i^2$, and $M_i$. Because the actual $M_i^{-1}$ appears in each term and every $M_i\geq M$, unequal allocation is retained explicitly and both contributions vanish as $O(M^{-1})$.

\paragraph{Recovering the main-paper bias guarantee.}
To obtain the proposition in the main paper, take $L\to\infty$ and $M=\min_iM_i\to\infty$. We assume target-support coverage, bounded $f$, suitable bounded higher-order and inverse-denominator moments, a permutation-equivariant pilot allocation, conditionally independent fresh batches, and the relevant laws of large numbers and uniform integrability. Zero-denominator and numerical-fallback probabilities must be $o(L^{-1})$. Under these regularity conditions, Eqs.~\plaineqref{eq:leading_ratio_bias}--\plaineqref{eq:unequal_rollout_moments} give
$L(\mathbb E[\widetilde T_k(f)]-T_{k,H}(f))\to0$, which is Eq.~\plaineqref{eq:main_leading_bias_proposition} and proves Proposition~\ref{prop:main_leading_bias} in the main paper.

The correction removes only the leading ratio-bias term. As a consistency check, when $f\equiv1$, $\overline X_f=\overline Y$ and $s_{XY}=s_Y^2$, so the correction is zero. Coordinate-wise correction of candidate indicators therefore sums to 1 in exact arithmetic; negative or nonfinite coordinates use the fallback in Appendix~\ref{app:validity}.

\section{Handling Failed Rollouts and Numerical Edge Cases}
\label{app:validity}

To keep the sampler defined for every outcome, we assign small positive weight to failed rollouts, correct any modified rollout proposal, and specify numerical fallbacks. We also quantify how failure smoothing changes the valid-only target.

\paragraph{Preventing failed rollouts from receiving zero weight.}
To prevent failed rollouts from creating zero denominators or unstable weights, we define a bounded, strictly positive complete-trajectory reward. For a valid, renderable SVG program $Y$, we use
\begin{equation}
    r_\beta(Y,q)
    =\exp\{\beta s(\mathcal R(Y),q)\}>0.
    \label{eq:positive_render_reward}
\end{equation}
where $\mathcal R(Y)$ is the final rendering, $s(\mathcal R(Y),q)$ is its bounded score for prompt $q$, and $\beta\geq0$ controls the strength of visual reweighting. The theory does not require a particular scorer architecture; it only requires the evaluator and its calibration to be fixed during inference.

For a malformed, timeout, overflow, or renderer-failure outcome, a valid visual score is unavailable. Instead of assigning zero reward, we set
\begin{equation}
    r_\beta(Y,q)=\epsilon,
    \qquad 0<\epsilon\ll1,
    \label{eq:failure_reward}
\end{equation}
where $\epsilon$ is a fixed small positive constant. This choice keeps every capped outcome in the probability space with nonzero reward while strongly downweighting failures.

\paragraph{Showing how failure smoothing changes the target.}
To make the difference between the ideal valid-only target and the implemented target explicit, let $\mathcal Y_q$ be the valid complete-program set and let $\mathcal F_q$ contain all capped failure outcomes. Their union $\overline{\mathcal Y}_q=\mathcal Y_q\cup\mathcal F_q$ is the operational outcome space. Define
\begin{equation}
    \Pi_{\alpha,\beta}^{\epsilon}(Y\mid q)
    \propto
    \overline p_\theta(Y\mid q)^\alpha r_\beta(Y,q),
    \qquad Y\in\overline{\mathcal Y}_q.
    \label{eq:operational_target}
\end{equation}
Here, \(\overline p_\theta\) is the probability distribution induced by running the base model under the operational caps. A rollout that produces a valid, fully terminated SVG before reaching any cap is assigned the probability of its generated token sequence. A rollout that reaches a cap or enters an unrecoverable malformed state is not discarded; its probability is assigned to the failure outcome \(\bot\). Thus, \(\overline p_\theta\) remains a normalized distribution over both valid completions and failed rollouts. The factor $\overline p_\theta(Y\mid q)^\alpha$ applies power sharpening, and $r_\beta(Y,q)$ supplies either the valid visual reward or the failure reward $\epsilon$. Thus, Eq.~\plaineqref{eq:operational_target} is the exact target implemented by the capped decoder before normalization.

To quantify how much target probability is assigned to failures, let $Z_{\mathrm{val}}$ and $Z_{\mathrm{fail}}$ be the unnormalized mass of Eq.~\plaineqref{eq:operational_target} on $\mathcal Y_q$ and $\mathcal F_q$, respectively. Then
\begin{equation}
    \delta_\epsilon
    =\frac{Z_{\mathrm{fail}}}{Z_{\mathrm{val}}+Z_{\mathrm{fail}}},
    \qquad
    Z_{\mathrm{fail}}
    =\epsilon\sum_{Y\in\mathcal F_q}
    \overline p_\theta(Y\mid q)^\alpha.
    \label{eq:operational_failure_probability}
\end{equation}
Here, $\delta_\epsilon$ is the operational failure probability, and $Z_{\mathrm{fail}}$ shows that failure mass is proportional to $\epsilon$. Conditioning the operational target on $\mathcal Y_q$ exactly recovers the valid-only target. Equivalently, after extending the valid-only target by zero mass on $\mathcal F_q$, its total-variation distance from the operational target is $\delta_\epsilon$, which vanishes as $\epsilon\to0$.

\paragraph{Keeping failed futures in the future-value expectation.}
To use the same future-value formula for both valid and failed continuations, we retain the telescoping visual product from Appendix~\ref{app:exact_conditional} for a valid suffix $C$. For a failure suffix $C=\bot$, we define $\Phi_{>k}^{\epsilon}(hb,\bot,q)>0$ so that the reward already accumulated by prefix $hb$, multiplied by this future factor, equals the fixed complete failure reward $\epsilon$. The future value then has the uniform form
\begin{equation}
\begin{aligned}
    V_k^{\epsilon}(hb,q)
    &=\mathbb E_{C\sim\overline p_\theta(\cdot\mid hb,q)}
    \!\left[
    \overline p_\theta(C\mid hb,q)^{\alpha-1}
    \Phi_{>k}^{\epsilon}(hb,C,q)
    \right],\\
    \Phi_{>k}^{\epsilon}&>0.
\end{aligned}
    \label{eq:positive_future_factor}
\end{equation}
Here, $\overline p_\theta(C\mid hb,q)$ is the capped probability of future outcome $C$, the exponent $\alpha-1$ supplies the likelihood power not already absorbed by sampling, and $\Phi_{>k}^{\epsilon}$ supplies the remaining visual or failure factor. The strict positivity statement ensures that the exact future value and its denominator remain defined. A malformed rollout is retained in the sample mean with a small contribution.

\paragraph{Preserving the target under modified rollout proposals.}
To preserve the same future-value target when render-guided pruning, early rejection, or other efficiency mechanisms change the rollout distribution, we let $q_{\mathrm{roll}}(C\mid hb,q)$ denote the actual rollout proposal and use
\begin{equation}
    V_k^{\epsilon}(hb,q)
    =\mathbb E_{C\sim q_{\mathrm{roll}}(\cdot\mid hb,q)}
    \!\left[
    \frac{\overline p_\theta(C\mid hb,q)^\alpha}
    {q_{\mathrm{roll}}(C\mid hb,q)}
    \Phi_{>k}^{\epsilon}(hb,C,q)
    \right].
    \label{eq:alternative_rollout_proposal}
\end{equation}
The numerator is the desired unnormalized contribution of $C$, and division by the actual proposal probability corrects its sampling frequency. The proposal must cover every positive-target outcome. When $q_{\mathrm{roll}}=\overline p_\theta$, one factor cancels and the expression reduces to Eq.~\plaineqref{eq:positive_future_factor}. In particular, averaging only render-filter survivors without this correction would estimate a different, survival-conditioned target.

\paragraph{Keeping selection valid when numerical correction fails.}
To keep selection defined when finite-precision arithmetic or the signed ratio correction fails, we distinguish three events. Let $\mathcal E_0$ be the event that the estimated total weight is zero or nonfinite. Let $\mathcal E_{\mathrm{corr}}$ be the event that the corrected probability vector has a negative coordinate, and let $\mathcal E_{\mathrm{num}}$ be the event that the correction itself is numerically nonfinite. For a bounded candidate statistic $f$, the implemented estimator is
\begin{equation}
\begin{aligned}
    T_{\mathrm{alg}}(f)
    &=\overline f\,\mathbf1[\mathcal E_0]
    +\widehat T_k^+(f)\,
    \mathbf1[\mathcal E_0^c\cap(\mathcal E_{\mathrm{corr}}\cup\mathcal E_{\mathrm{num}})]\\
    &\quad+\widetilde T_k(f)\,
    \mathbf1[\mathcal E_0^c\cap\mathcal E_{\mathrm{corr}}^c\cap\mathcal E_{\mathrm{num}}^c],
\end{aligned}
\label{eq:fallback_statistic}
\end{equation}
Here, $\mathbf1[\cdot]$ is an event indicator, the superscript $c$ denotes the complement of an event, and $\overline f=L^{-1}\sum_i f(B_i)$ is the statistic under a uniform distribution over the $L$ parser-valid particles. Eq.~\plaineqref{eq:fallback_statistic} implements three branches. If $\mathcal E_0$ occurs, selection is uniform over valid particles. If the total weight is valid but the correction is negative or nonfinite, the method uses the uncorrected self-normalized estimator $\widehat T_k^+$. Otherwise, it uses the corrected estimator $\widetilde T_k$ and renormalizes the corrected vector in floating-point arithmetic.

\paragraph{Separating fallback error from estimator error.}
To separate the usual corrected-estimator error from the additional error caused by fallbacks, we define the numerically valid event $\mathcal G=\mathcal E_0^c\cap\mathcal E_{\mathrm{num}}^c$ and let $\|f\|_\infty=\sup_b|f(b)|$. Because this section makes failure smoothing explicit, we write the finite-horizon reference target as $T_{k,H}^{\epsilon}$; the algorithmic estimators below are computed from the same $\epsilon$-smoothed weights. Then
\begin{equation}
\begin{aligned}
    |\mathbb E[T_{\mathrm{alg}}(f)-T_{k,H}^{\epsilon}(f)]|
    &\leq
    \left|\mathbb E[(\widetilde T_k(f)-T_{k,H}^{\epsilon}(f))
    \mathbf1[\mathcal G]]\right|\\
    &\quad+2\|f\|_\infty
    \{\Pr(\mathcal E_0)+\Pr(\mathcal E_{\mathrm{num}})\}\\
    &\quad+
    \left(\mathbb E|\widehat T_k^+(f)-\widetilde T_k(f)|^2\right)^{1/2}\\
    &\qquad\times
    \Pr(\mathcal E_{\mathrm{corr}})^{1/2}.
\end{aligned}
\label{eq:fallback_error_bound}
\end{equation}
The three terms respectively cover the corrected estimator on valid numerical outcomes, zero or nonfinite weights, and fallback caused by a negative corrected coordinate. Thus, the leading-bias result applies on the regular event, while fallback frequencies remain a separate implementation diagnostic.

\paragraph{Keeping weights stable and visual scores comparable.}
To avoid numerical underflow, all candidate likelihoods, suffix likelihoods, visual factors, and importance weights are accumulated in log space and normalized with log-sum-exp. Padding tokens are excluded from both the likelihood and the model context.

To keep scores comparable, current candidates and future completions use the same canonicalization, viewport, background, raster resolution, parser, and renderer. Caching and batching do not remove generated tokens, renderer calls, or evaluator calls from the reported cost. Because the visual factors telescope, only the horizon-endpoint visual score is needed for the rollout weight; intermediate parsing still detects boundaries and failures.

\section{Complete CANVAS Procedure and Sources of Approximation}
\label{app:procedure}

To provide an end-to-end specification of CANVAS, we assemble the preceding components into one navigation procedure, clarify which guarantees are exact and which rely on approximation, and state the shared experimental configuration.

\paragraph{Generating one SVG end to end.}
To generate one SVG, CANVAS receives prompt $q$, candidate count $L$, pilot count $m_0$, rollout horizon $H$, likelihood-sharpening parameter $\alpha$, visual-weight parameter $\beta$, failure reward $\epsilon$, and a rollout-and-rendering budget. At decision step $k$, let $h=H_{k-1}$ be the current committed prefix. CANVAS repeats the following procedure at every parser-detected stroke boundary.

\paragraph{Step 1: Candidate generation.}
To approximate the full next-stroke candidate space, independently sample $L$ parser-complete candidate strokes from the actual valid proposal $q_k(\cdot\mid h,q)$. For each particle, compute the importance factor in Eq.~\plaineqref{eq:appendix_candidate_importance}. Repeated token-identical candidates remain separate particles because they arose from separate proposal draws.

\paragraph{Step 2: Estimate where fresh rollouts are needed.}
To estimate how many fresh rollouts each candidate should receive, generate $m_0$ capped pilot futures for every particle. Use Eq.~\plaineqref{eq:pilot_diagnosis} to estimate its future value $\widehat V_i^{(0)}$, rollout variance $\widehat\sigma_i^2$, and per-rollout cost $\widehat c_i$. A malformed or unrenderable pilot is retained and receives the failure reward $\epsilon$ from Eq.~\plaineqref{eq:failure_reward}.

\paragraph{Step 3: Budget allocation.}
To spend the fresh-rollout budget where it most reduces decision error, use Eq.~\plaineqref{eq:pilot_plugin_coefficients} to estimate each candidate's decision-error coefficient. Determine the integer rollout count $M_i$ under the lower, upper, and total-budget constraints using Eqs.~\plaineqref{eq:clipped_kkt_allocation} and~\plaineqref{eq:rounding_marginal_gain}. Freeze every $M_i$ before any fresh outcome is generated.

\paragraph{Step 4: Estimate future value with independent rollouts.}
To obtain the final conditionally unbiased finite-horizon estimate, generate $M_i$ new capped futures for candidate $B_i$, independently of its pilots, and compute $\widehat V_{i,H}$ using Eq.~\plaineqref{eq:operational_rollout_unbiasedness}. Pilot outcomes are not reused in this estimate.

\paragraph{Step 5: Combine current and future evidence.}
To combine the current-stroke factor with its estimated future value, compute the unnormalized particle mass $Y_i=A_i\widehat V_{i,H}$. Normalize these masses using Eq.~\plaineqref{eq:normalized_selection} in the main paper to obtain the uncorrected particle distribution. Eq.~\plaineqref{eq:unnormalized_unbiasedness} guarantees unbiasedness of the numerator and denominator masses before this normalization.

\paragraph{Step 6: Reduce normalization bias.}
To reduce the leading bias introduced by the random normalizing denominator, compute the particle variance and covariance statistics in Eqs.~\plaineqref{eq:ratio_particle_moments} and~\plaineqref{eq:ratio_sample_moments}. Apply Eq.~\plaineqref{eq:main_analytic_ratio_correction} in the main paper to the indicator of each candidate. The resulting corrected coordinates form the signed vector $\widetilde{\boldsymbol\rho}$.

\paragraph{Step 7: Fallback and commit.}
To ensure that the committed choice always comes from a valid probability distribution, select the corrected, uncorrected, or uniform valid-particle distribution according to Eq.~\plaineqref{eq:fallback_statistic}. Sample the next stroke from that distribution and append it to the committed prefix. If an operational program cap is exhausted or the valid current-candidate proposal has zero mass, record a capped-decoding failure rather than committing a malformed segment.

\paragraph{Step 8: Termination.}
To complete generation, repeat Steps 1--7 until the parser observes that all remaining SVG/XML tags are closed and EOS is generated. Return the resulting complete SVG program. Every committed prefix remains parser-valid; malformed or unrenderable hypothetical futures influence the lookahead estimate only through their small $\epsilon$ reward and are never committed directly.

\paragraph{Clarifying what is exact and what is approximate.}
To interpret the guarantees correctly, separate four levels. The complete-SVG target and its boundary-wise marginalization are exact. With fixed horizon and proposal support, candidate importance averages and independent fresh-rollout means are unbiased before normalization. Random normalization introduces ratio bias, for which Appendix~\ref{app:ratio_bias} removes only the leading asymptotic term. Finite $H$, restricted support, $\epsilon$-smoothing, integer allocation, and fallback are separate approximations. The visual evaluator also defines the reward being optimized; matching that target does not guarantee unmeasured human preference.

\paragraph{Explaining why correction and allocation are both needed.}
To clarify why both mechanisms are needed, the ratio correction reduces systematic bias from random normalization, whereas the adaptive counts $M_i$ reduce finite-budget variance. Eq.~\plaineqref{eq:main_uniform_adaptive_comparison} establishes the allocation result only for the stated first-order decision-error surrogate; downstream metrics also depend on reward fidelity, horizon $H$, and proposal coverage.

\paragraph{Keeping backbone comparisons reproducible.}
To make comparisons across backbones reproducible, the main experiments use four NVIDIA H200 GPUs and set $\alpha=2$. On HeisenVec, we evaluate six representative autoregressive SVG backbones: IconShop~\citep{wu2023iconshop}, LLM4SVG~\citep{xing2025llm4svg}, CodeLLaMA-7B, vHector-3B, OmniSVG~\citep{yang2025omnisvg}, and vHector-8B~\citep{zini2025vhector}. On the unified evaluation set introduced by IntroSVG, we report the three backbones in its main comparison, OmniSVG, SVGen, and IntroSVG.

For every backbone, all model parameters remain frozen; CANVAS changes only the inference procedure and performs no fine-tuning or gradient update. Each backbone retains its native tokenizer, prompt format, and SVG serialization. Stroke boundaries are detected online by the incremental parser in Sec.~\ref{sec:stroke_framework}, while benchmark prompts, rendering settings are all following original settings from both benchmarks.

\section{Additional Ablation Studies}
\label{app:ablation_studies}

In this section, we present additional ablation studies of the main design choices in CANVAS. Unless noted otherwise, all experiments use vHector-8B and the HeisenVec evaluation protocol from Tab.~\ref{tab:heisenvec_main} in the main paper. \textbf{Ours} denotes the complete CANVAS configuration with $H=1$ and $\alpha=2$. Arrows in the table headers indicate the preferred metric direction, and bold marks the best result in each table.

\subsection{Impact of the Rollout Horizon}
\label{app:horizon_ablation}

In our framework, the rollout horizon $H$ limits how many future strokes are evaluated after each current candidate, and we use $H=1$ by default. To evaluate the impact of this choice, we compare three variants. \textbf{H1(Ours)}, \textbf{H2}, and \textbf{H3} allow at most one, two, and three additional parser-complete future strokes, respectively, while \textbf{HEOS} continues to valid EOS subject to the operational caps in Appendix~\ref{app:boundary_rule}. Every variant stops earlier if valid EOS is reached. All reported runtimes were measured using four NVIDIA H200 GPUs. As shown in Tab.~\ref{tab:rollout_horizon_ablation}, H1 achieves the best performance on most metrics while requiring only about $17.8$ seconds per sample. Deeper rollouts improve a few individual metrics, but at a substantial runtime cost. These results show that H1 provides the strongest overall quality--efficiency balance, supporting its use as our default horizon.

\begin{table*}[t]
\centering
\caption{Evaluation on the rollout horizon and inference time.}
\label{tab:rollout_horizon_ablation}
\vspace{-0.5mm}
\suppTableSetup
\begin{tabular*}{\textwidth}{@{\extracolsep{\fill}}lrrrrrrrrrrrrr@{}}
\toprule
& \multicolumn{6}{c}{Benchmark metrics}
& \multicolumn{6}{c}{Consistency metrics}
& \\
\cmidrule(lr){2-7}\cmidrule(lr){8-13}
Variant
& CLIP-S$\uparrow$
& FID$\downarrow$
& HPSv2$\uparrow$
& SSIM$\uparrow$
& LPIPS$\downarrow$
& DINO-S$\uparrow$
& LCI$_{9\times9}\uparrow$
& CVIU-C$\uparrow$
& SC$\uparrow$
& SL$\uparrow$
& CE$\uparrow$
& LaDe$\uparrow$
& \shortstack{Avg. time/sample (s)\\4$\times$H200$\downarrow$}
\\
\midrule
H1(Ours)
& \textbf{72.324}
& 151.634
& \textbf{20.219}
& 59.516
& 56.073
& 59.400
& 0.7736
& 0.5256
& \textbf{0.9312}
& \textbf{0.9274}
& \textbf{0.9159}
& \textbf{4.7362}
& \textbf{$\approx$17.8\,s}
\\
H2
& 69.977
& \textbf{102.383}
& 15.806
& 62.604
& 59.617
& 56.048
& 0.7976
& 0.5811
& 0.5698
& 0.4184
& 0.6177
& 3.1752
& $\approx$38.2\,s
\\
H3
& 71.046
& 198.614
& 17.844
& \textbf{63.681}
& \textbf{54.196}
& \textbf{65.620}
& \textbf{0.8377}
& 0.6180
& 0.6951
& 0.5182
& 0.7338
& 3.6844
& $\approx$63.4\,s
\\
HEOS
& 70.358
& 205.196
& 17.300
& 61.466
& 55.391
& 59.878
& 0.8270
& \textbf{0.6212}
& 0.6456
& 0.4816
& 0.6971
& 3.5956
& $\approx$227.5\,s
\\
\bottomrule
\end{tabular*}
\vspace{-1mm}
\end{table*}

\subsection{Evaluation on Decoding and Sampling Alternatives}
\label{app:decoding_ablation}

\begin{figure*}[t]
\centering
\includegraphics[width=\textwidth]{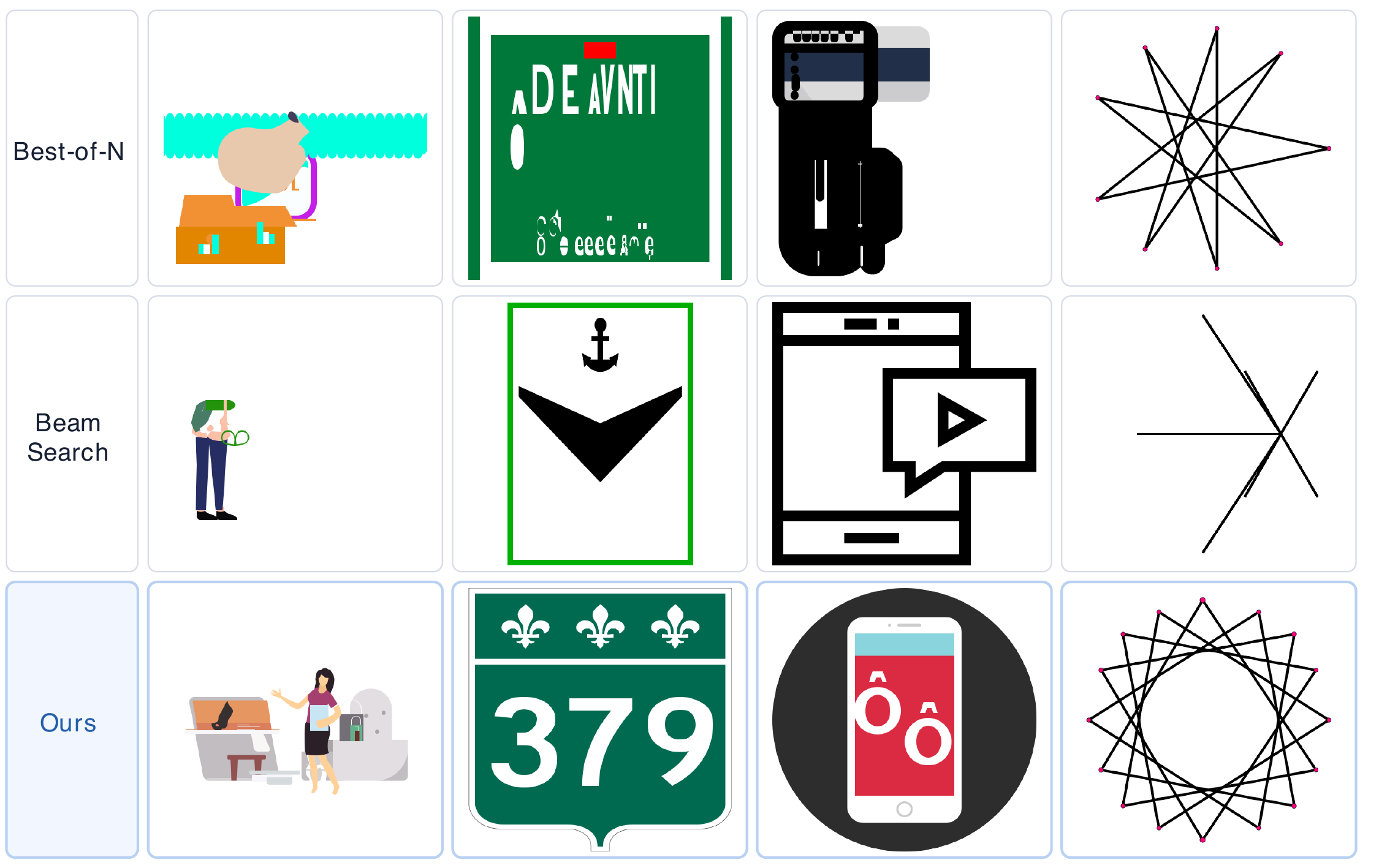}
\caption{\textbf{Qualitative evaluation on decoding and sampling alternatives.} Rows compare Best-of-$N$ ($N=5$), beam search ($B=5$), and CANVAS (Ours, H1) using the same vHector-8B backbone and sample IDs. From left to right, the prompts describe a laboratory scene, a traffic sign, a mobile phone, and a geometric star pattern. CANVAS more consistently preserves the requested content and global structure.}
\label{fig:decoding_qualitative}
\vspace{-2mm}
\end{figure*}

In our framework, CANVAS uses parser-aligned, future-aware decisions rather than a conventional sequence decoder. To determine whether its gains could instead come from a generic decoding or sampling strategy, we compare it with native decoding, four sampling temperatures, beam search with width $B=5$, and Best-of-$N$ selection with $N=5$, all using the same vHector-8B backbone. As shown in Tab.~\ref{tab:decoding_ablation}, temperature scaling and beam search improve only isolated metrics. Best-of-$N$ is the strongest conventional alternative, but CANVAS improves eight of its 12 metrics and also achieves the best overall result on eight metrics. This comparison shows that CANVAS's gains do not follow simply from sharpening the decoder, maintaining a beam, or selecting among more complete samples. Figure~\ref{fig:decoding_qualitative} provides a qualitative comparison on the same sample IDs.

\begin{table*}[t]
\centering
\caption{Evaluation on decoding and sampling alternatives.}
\label{tab:decoding_ablation}
\vspace{-0.5mm}
\suppTableSetup
\begin{tabular*}{\textwidth}{@{\extracolsep{\fill}}lrrrrrrrrrrrr@{}}
\toprule
& \multicolumn{6}{c}{Benchmark metrics}
& \multicolumn{6}{c}{Consistency metrics}
\\
\cmidrule(lr){2-7}\cmidrule(lr){8-13}
Variant
& CLIP-S$\uparrow$
& FID$\downarrow$
& HPSv2$\uparrow$
& SSIM$\uparrow$
& LPIPS$\downarrow$
& DINO-S$\uparrow$
& LCI$_{9\times9}\uparrow$
& CVIU-C$\uparrow$
& SC$\uparrow$
& SL$\uparrow$
& CE$\uparrow$
& LaDe$\uparrow$
\\
\midrule
Native decoding
& 66.787 & 117.741 & 15.213 & 66.229 & 57.599 & 49.488
& 0.6818 & 0.4535 & 0.5148 & 0.3436 & 0.4816 & 3.0000 \\
Temperature ($T=1$)
& 67.966 & 103.390 & 14.987 & 64.262 & 59.305 & 52.412
& 0.7542 & 0.5198 & 0.4897 & 0.3463 & 0.5288 & 2.8505 \\
Temperature ($T=0.5$)
& 67.252 & 110.003 & 15.144 & 65.867 & 57.643 & 51.360
& 0.7256 & 0.4810 & 0.5359 & 0.3579 & 0.5188 & 3.0506 \\
Temperature ($T=0.2$)
& 66.893 & 116.173 & 15.202 & 66.083 & 57.750 & 49.867
& 0.6896 & 0.4570 & 0.5239 & 0.3490 & 0.4896 & 3.0684 \\
Temperature ($T=0.1$)
& 66.923 & 117.194 & 15.217 & 66.221 & 57.597 & 49.766
& 0.6887 & 0.4547 & 0.5172 & 0.3470 & 0.4828 & 3.0075 \\
Beam search ($B=5$)
& 66.380 & 144.842 & 15.452 & \textbf{69.358} & 56.884 & 46.081
& 0.6115 & 0.3969 & 0.4959 & 0.3236 & 0.4771 & 3.0233 \\
Best-of-$N$ ($N=5$)
& 68.582 & \textbf{102.333} & 15.734 & 64.739 & 57.978 & 56.817
& \textbf{0.7965} & \textbf{0.5622} & 0.6076 & 0.4386 & 0.6448 & 3.3594 \\
\midrule
\textbf{Ours}
& \textbf{72.324} & 151.634 & \textbf{20.219} & 59.516 & \textbf{56.073} & \textbf{59.400}
& 0.7736 & 0.5256 & \textbf{0.9312} & \textbf{0.9274} & \textbf{0.9159} & \textbf{4.7362} \\
\bottomrule
\end{tabular*}
\vspace{-1mm}
\end{table*}

\subsection{Impact of Power Sharpening and Rendered Visual Weighting}
\label{app:power_ablation}

In our framework, power sharpening increases the relative weight of trajectories preferred by the base model, while rendered visual weighting favors complete SVGs that match the prompt. To evaluate the contribution of these two components, we compare native decoding, power-only sampling, and full CANVAS. As shown in Tab.~\ref{tab:power_weighting_ablation}, power-only sampling improves eight of the 12 metrics over native decoding, but degrades FID, HPSv2, SSIM, and LPIPS. Adding rendered visual weighting improves 11 of the 12 metrics over power-only sampling and gives the best overall result on 10 metrics. These results support using the two signals together: power sharpening preserves the base model's trajectory preference, while rendered feedback guides selection toward visually stronger complete SVGs.

\begin{table*}[t]
\centering
\caption{Evaluation on power sharpening and rendered visual weighting.}
\label{tab:power_weighting_ablation}
\vspace{-0.5mm}
\suppTableSetup
\begin{tabular*}{\textwidth}{@{\extracolsep{\fill}}lrrrrrrrrrrrr@{}}
\toprule
& \multicolumn{6}{c}{Benchmark metrics}
& \multicolumn{6}{c}{Consistency metrics}
\\
\cmidrule(lr){2-7}\cmidrule(lr){8-13}
Variant
& CLIP-S$\uparrow$
& FID$\downarrow$
& HPSv2$\uparrow$
& SSIM$\uparrow$
& LPIPS$\downarrow$
& DINO-S$\uparrow$
& LCI$_{9\times9}\uparrow$
& CVIU-C$\uparrow$
& SC$\uparrow$
& SL$\uparrow$
& CE$\uparrow$
& LaDe$\uparrow$
\\
\midrule
Power only
& 67.116 & 174.587 & 15.207 & 66.054 & 57.667 & 50.209
& 0.7031 & 0.4636 & 0.5247 & 0.3507 & 0.4978 & 3.0230 \\
Native decoding
& 66.787 & \textbf{117.741} & 15.213 & \textbf{66.229} & 57.599 & 49.488
& 0.6818 & 0.4535 & 0.5148 & 0.3436 & 0.4816 & 3.0000 \\
Full CANVAS(Ours)
& \textbf{72.324} & 151.634 & \textbf{20.219} & 59.516 & \textbf{56.073} & \textbf{59.400}
& \textbf{0.7736} & \textbf{0.5256} & \textbf{0.9312} & \textbf{0.9274} & \textbf{0.9159} & \textbf{4.7362} \\
\bottomrule
\end{tabular*}
\vspace{-1mm}
\end{table*}

\subsection{\texorpdfstring{Impact of the Power Coefficient $\alpha$}{Impact of the Power Coefficient alpha}}
\label{app:power_sweep_ablation}

In our framework, the coefficient $\alpha$ controls how strongly the complete-sequence likelihood is sharpened, and we use $\alpha=2$ by default. To evaluate this choice, we test $\alpha\in\{1,2,3,5,10\}$ while keeping the remaining CANVAS components fixed. As shown in Tab.~\ref{tab:power_sweep_ablation}, $\alpha=1$ gives the best FID and SSIM, while $\alpha=3$ gives the best CLIP-S, DINO-S, and CVIU-C. The default $\alpha=2$ leads the other seven metrics, including all four global-consistency scores SC, SL, CE, and LaDe. It therefore provides the strongest overall balance for the behavior targeted by CANVAS.

\begin{table*}[t]
\centering
\caption{Evaluation on the power coefficient $\alpha$.}
\label{tab:power_sweep_ablation}
\vspace{-0.5mm}
\suppTableSetup
\begin{tabular*}{\textwidth}{@{\extracolsep{\fill}}lrrrrrrrrrrrr@{}}
\toprule
& \multicolumn{6}{c}{Benchmark metrics}
& \multicolumn{6}{c}{Consistency metrics}
\\
\cmidrule(lr){2-7}\cmidrule(lr){8-13}
$\alpha$
& CLIP-S$\uparrow$
& FID$\downarrow$
& HPSv2$\uparrow$
& SSIM$\uparrow$
& LPIPS$\downarrow$
& DINO-S$\uparrow$
& LCI$_{9\times9}\uparrow$
& CVIU-C$\uparrow$
& SC$\uparrow$
& SL$\uparrow$
& CE$\uparrow$
& LaDe$\uparrow$
\\
\midrule
$1$
& 69.961 & \textbf{103.224} & 15.799 & \textbf{66.666} & 56.861 & 57.000
& 0.7543 & 0.5212 & 0.6370 & 0.4841 & 0.6337 & 3.3971 \\
$2$ (Ours)
& 72.324 & 151.634 & \textbf{20.219} & 59.516 & \textbf{56.073} & 59.400
& \textbf{0.7736} & 0.5256 & \textbf{0.9312} & \textbf{0.9274} & \textbf{0.9159} & \textbf{4.7362} \\
$3$
& \textbf{74.045} & 143.156 & 18.278 & 54.977 & 59.161 & \textbf{61.718}
& 0.6924 & \textbf{0.5401} & 0.8396 & 0.6892 & 0.8094 & 4.4342 \\
$5$
& 71.066 & 142.134 & 18.420 & 57.578 & 57.494 & 57.055
& 0.7062 & 0.4452 & 0.8145 & 0.6979 & 0.6661 & 4.0851 \\
$10$
& 72.122 & 139.098 & 19.051 & 58.394 & 56.643 & 57.541
& 0.7095 & 0.4420 & 0.8365 & 0.6805 & 0.7462 & 4.4342 \\
\bottomrule
\end{tabular*}
\vspace{-1mm}
\end{table*}

\subsection{Impact of Adaptive Allocation and Estimator Reliability}
\label{app:allocation_reliability_ablation}

In our framework, rollout counts depend jointly on estimated uncertainty, influence on the final decision, and rollout cost. Pilot rollouts determine these counts, fresh rollouts estimate candidate values, and an analytic correction reduces the leading bias from random normalization. To evaluate these choices, we replace the allocation with uniform or variance-only allocation, remove its cost term, reuse pilot outcomes, remove the ratio correction, or remove both the pilot/fresh split and the correction. As shown in Tab.~\ref{tab:allocation_reliability_ablation}, individual variants lead on isolated metrics, but the complete estimator achieves the overall best performance. This result supports combining influence- and cost-aware allocation with fresh evaluation and ratio correction, particularly for global consistency.

\begin{table*}[t]
\centering
\caption{Evaluation on adaptive allocation and estimator-reliability components.}
\label{tab:allocation_reliability_ablation}
\vspace{-0.5mm}
\suppTableSetup
\begin{tabular*}{\textwidth}{@{\extracolsep{\fill}}lrrrrrrrrrrrr@{}}
\toprule
& \multicolumn{6}{c}{Benchmark metrics}
& \multicolumn{6}{c}{Consistency metrics}
\\
\cmidrule(lr){2-7}\cmidrule(lr){8-13}
Variant
& CLIP-S$\uparrow$
& FID$\downarrow$
& HPSv2$\uparrow$
& SSIM$\uparrow$
& LPIPS$\downarrow$
& DINO-S$\uparrow$
& LCI$_{9\times9}\uparrow$
& CVIU-C$\uparrow$
& SC$\uparrow$
& SL$\uparrow$
& CE$\uparrow$
& LaDe$\uparrow$
\\
\midrule
Uniform allocation
& 72.913 & 135.835 & 19.080 & 58.664 & \textbf{55.298} & 60.148
& 0.6721 & 0.4705 & 0.8909 & 0.7479 & 0.8618 & 4.5601 \\
Variance-only allocation
& 71.651 & 140.309 & 19.511 & \textbf{62.616} & 57.320 & \textbf{67.808}
& 0.6637 & \textbf{0.5588} & 0.8575 & 0.7600 & 0.8814 & 4.3685 \\
Influence without cost
& 73.173 & 139.924 & 17.332 & 54.226 & 61.401 & 51.610
& 0.6919 & 0.4426 & 0.7591 & 0.6400 & 0.7244 & 3.9695 \\
Reuse pilot outcomes
& \textbf{74.017} & \textbf{135.622} & 18.926 & 54.611 & 58.873 & 61.841
& 0.7215 & 0.5082 & 0.8046 & 0.7479 & 0.8499 & 4.0851 \\
No ratio correction
& 72.859 & 136.713 & 18.657 & 54.807 & 62.281 & 63.613
& 0.6588 & 0.4631 & 0.8046 & 0.7520 & 0.7718 & 4.0851 \\
No split or correction
& 73.423 & 136.300 & 17.969 & 52.747 & 61.687 & 56.415
& 0.7038 & 0.5077 & 0.7700 & 0.6979 & 0.7156 & 3.9297 \\
\midrule
\textbf{Ours}
& 72.324 & 151.634 & \textbf{20.219} & 59.516 & 56.073 & 59.400
& \textbf{0.7736} & 0.5256 & \textbf{0.9312} & \textbf{0.9274} & \textbf{0.9159} & \textbf{4.7362} \\
\bottomrule
\end{tabular*}
\vspace{-1mm}
\end{table*}

\subsection{Impact of the Decision Unit}
\label{app:decision_reward_ablation}

In our framework, parser-detected stroke boundaries define the decision unit. To evaluate this choice, we compare stroke-wise decisions with token-wise decisions and fixed-length blocks of $m=20$ tokens. As shown in In Tab.~\ref{tab:decision_reward_ablation}, token-wise and fixed-length decisions achieve the best results on a few metrics but incur substantial additional runtime. Stroke-wise CANVAS takes approximately $17.8$ seconds and achieves the best result on most of the metrics. Fixed boundaries may also split a coordinate, path command, or XML attribute. These results support parser-detected strokes as a meaningful and efficient decision unit.

\begin{table*}[t]
\centering
\caption{Evaluation on token-wise, fixed-length, and stroke-wise decision units.}
\label{tab:decision_reward_ablation}
\vspace{-0.5mm}
\suppTableSetup
\begin{tabular*}{\textwidth}{@{\extracolsep{\fill}}lrrrrrrrrrrrrr@{}}
\toprule
& \multicolumn{6}{c}{Benchmark metrics}
& \multicolumn{6}{c}{Consistency metrics}
& \\
\cmidrule(lr){2-7}\cmidrule(lr){8-13}
Variant
& CLIP-S$\uparrow$
& FID$\downarrow$
& HPSv2$\uparrow$
& SSIM$\uparrow$
& LPIPS$\downarrow$
& DINO-S$\uparrow$
& LCI$_{9\times9}\uparrow$
& CVIU-C$\uparrow$
& SC$\uparrow$
& SL$\uparrow$
& CE$\uparrow$
& LaDe$\uparrow$
& \shortstack{Avg. time/sample (s)\\4$\times$H200$\downarrow$}
\\
\midrule
Token-wise decision
& 68.878 & \textbf{66.574} & 13.753 & \textbf{69.913} & 62.766 & 32.981
& \textbf{0.8801} & 0.3121 & 0.4723 & 0.2663 & 0.5230 & 2.6875
& $\approx$7{,}112.4\,s \\
Fixed block ($m=20$)
& 71.326 & 89.597 & 14.395 & 62.912 & 61.039 & 56.761
& 0.8195 & \textbf{0.6021} & 0.6355 & 0.3769 & 0.6657 & 3.7224
& $\approx$165.9\,s \\
\midrule
\textbf{Ours}
& \textbf{72.324} & 151.634 & \textbf{20.219} & 59.516 & \textbf{56.073} & \textbf{59.400}
& 0.7736 & 0.5256 & \textbf{0.9312} & \textbf{0.9274} & \textbf{0.9159} & \textbf{4.7362}
& \textbf{$\approx$17.8\,s} \\
\bottomrule
\end{tabular*}
\vspace{-1mm}
\end{table*}

\section{Benchmark Datasets and Official Evaluation Metrics}
\label{app:benchmark_metrics}

We retain the official evaluation protocols and metric directions of vHector/HeisenVec~\citep{zini2025vhector} and IntroSVG~\citep{wang2026introsvg}.

\paragraph{HeisenVec.}
HeisenVec is a million-scale, richly captioned SVG dataset spanning diverse visual styles and sequence lengths, with samples extending to 32k tokens, and is designed for long-context Text-to-SVG modeling~\citep{zini2025vhector}.
Following its official benchmark, we report six complementary measures.
CLIP Score (CLIP-S) is the cosine similarity between CLIP image and text embeddings and measures alignment between a rendered SVG and its source caption.
Fr\'echet Inception Distance (FID) compares the distributions of Inception-v3 features from generated and reference renders; lower values indicate closer distributional alignment.
Human Preference Score v2 (HPSv2) is a learned caption-image preference surrogate trained to approximate which output a human would favor.
Structural Similarity Index Measure (SSIM) compares local luminance, contrast, and structure between a generated render and its paired reference.
Learned Perceptual Image Patch Similarity (LPIPS) measures perceptual distance between the pair using learned deep image features.
DINO Similarity (DINO-S) compares their DINO feature representations and provides a complementary feature-level fidelity measure.
Higher values are preferred for CLIP-S, HPSv2, SSIM, and DINO-S, whereas lower values are preferred for FID and LPIPS.

\paragraph{IntroSVG.}
IntroSVG constructs a unified held-out evaluation set of 1,400 samples, stratified across the data sources used by LLM4SVG, OmniSVG, and SVGen and excluded from its training corpus~\citep{wang2026introsvg}.
Average token count (Avg. Token) is the length of the generated SVG after tokenization by the Qwen2.5 tokenizer and measures code conciseness.
Render Success Rate (RSR) is the percentage of generated SVG programs that CairoSVG renders successfully.
FID has the distributional interpretation given above.
CLIP text-to-image similarity (CLIP-T2I) measures semantic alignment between the prompt and rendered SVG through CLIP embeddings.
Aesthetic Score applies a pretrained image-aesthetic predictor to the render, while HPS applies a learned caption-image human-preference predictor.
Lower values are preferred for Avg. Token and FID, while higher values are preferred for RSR, CLIP-T2I, Aesthetic Score, and HPS.

\section{Definitions and Qualitative Interpretation of the Consistency Metrics}
\label{app:consistency_metric_visualizations}

To make the six additional consistency metrics in the main paper easy to interpret, this section identifies the source of each metric, gives its precise definition and dataset-level aggregation, and then explains what a higher score looks like in practice. The metrics measure consistency in two different ways. SC, SL, CE, and LaDe use visual-judging criteria adapted to our setting, whereas LCI$_{9\times9}$ and CVIU-C compute consistency directly from image edges using fixed deterministic rules. All six are higher-is-better. SC, SL, CE, LCI$_{9\times9}$, and CVIU-C lie in $[0,1]$, while LaDe lies in $[1,5]$.

\paragraph{Two complementary evaluation protocols.}
We first evaluate SC, SL, CE, and LaDe as VLM-judged metrics. SC, SL, and CE adapt the corresponding symbolic-visual rubrics from Latent Canvas~\citep{zheng2026latentcanvas}, while LaDe adapts the cross-layer-consistency stage of the layered-design judge in~\citet{lungustan2026lade}. For each sample and method variant, a single blinded GPT-5.5 call jointly returns all four scores in one JSON response. The judge receives the exact caption, the final SVG rendered on a white $512\times512$ canvas, SVG source statistics, and at most six pseudo-layers formed by grouping contiguous drawables in z-order. It is also shown a second rendering generated from the same caption solely to calibrate failure severity; this image is not ground truth and is not assumed to be better. Because these four metrics are defined by frozen VLM rubrics rather than closed-form pixel calculations, the corresponding subsections quote their metric-specific prompt clauses separately so that each definition can be read in place. This separation is only for presentation: all four scores are obtained together in the same judge call.

In contrast, LCI$_{9\times9}$ and CVIU-C are deterministic metrics computed from image edges by explicit formulas, without a VLM judge. We apply the same frozen raster-to-edge adapter to every method: the $512\times512$ white-background rendering is resized to $128\times128$, its RGB distance from white is normalized at the $99.5$th percentile, the result is smoothed with a Gaussian kernel of $\sigma=0.8$, and a Canny edge map is extracted with thresholds 32 and 96. The subsections below define how LCI$_{9\times9}$ and CVIU-C summarize different properties of this common edge map.

For any one of the six metrics, let $m_n$ be its score on evaluation sample $n$. The value reported in the tables is the arithmetic mean over the $N$ evaluated samples:
\begin{equation}
    \overline m=\frac{1}{N}\sum_{n=1}^{N}m_n.
    \label{eq:consistency_metric_dataset_average}
\end{equation}
For readability, table headers use the metric name without the overbar.

\paragraph{How to read the qualitative comparisons.}
Figures~\ref{fig:sc_metric_semantics}--\ref{fig:cviu_metric_semantics} show two same-prompt, same-sample-ID pairs for each metric. ``Higher'' and ``Lower'' compare only the two outputs within one pair; they do not denote dataset-wide maxima or minima. Every displayed SVG is render-valid and nonempty, and its content is shown unchanged on the same white canvas without cropping. The examples are chosen to make the property targeted by each metric visible.

\subsection{Structural Coherence (SC)}
\label{app:sc_metric_semantics}

Latent Canvas defines Structural Coherence as a $[0,1]$ VLM-judge score for whether lines are connected and shapes are closed and logically constructed~\citep{zheng2026latentcanvas}. We transfer this rubric from ASCII art to SVG while retaining the same score range.

\noindent\textbf{SC clause in the frozen joint judge prompt.}
\emph{``\texttt{latent\_structural\_coherence} $[0,1]$, adapted from Latent Canvas: are contours, lines and shapes connected/closed when they should be, stable, recognizable, and free of broken or disconnected construction?''}

This clause asks whether an SVG's local geometry forms stable, recognizable objects. It checks whether contours and parts join, close, and remain separate where the caption requires, and penalizes broken or unclosed contours, unintended gaps, accidentally fused parts, and unstable shape construction.

\begin{figure*}[p]
\centering
\setlength{\tabcolsep}{1.5pt}
\begin{tabular*}{\textwidth}{@{\extracolsep{\fill}}cccc@{}}
\multicolumn{2}{c}{\small\textbf{Sample 5756}}
& \multicolumn{2}{c}{\small\textbf{Sample 3546}} \\
{\scriptsize\textbf{Higher SC}}
& {\scriptsize\textbf{Lower SC}}
& {\scriptsize\textbf{Higher SC}}
& {\scriptsize\textbf{Lower SC}} \\[-0.5mm]
\includegraphics[width=.235\textwidth]{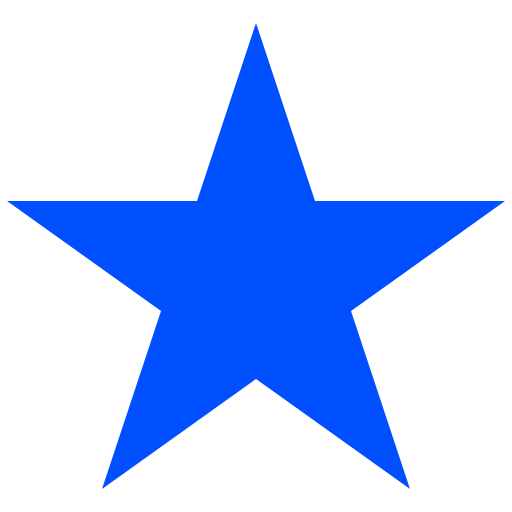}
& \includegraphics[width=.235\textwidth]{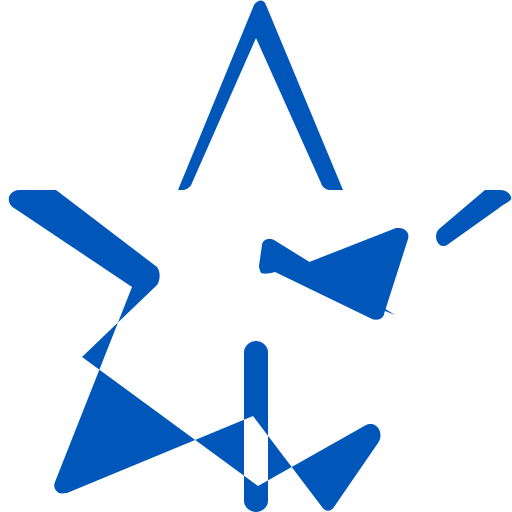}
& \includegraphics[width=.235\textwidth]{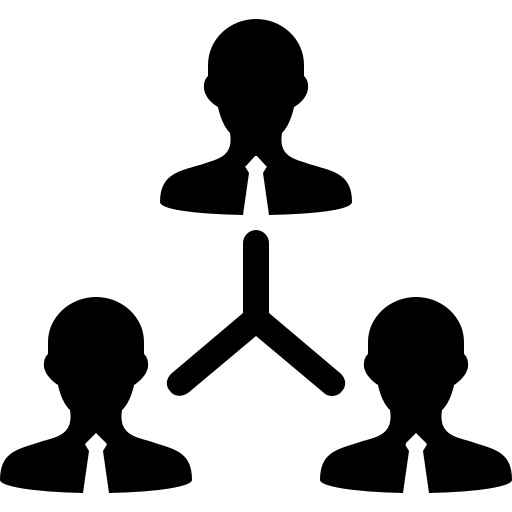}
& \includegraphics[width=.235\textwidth]{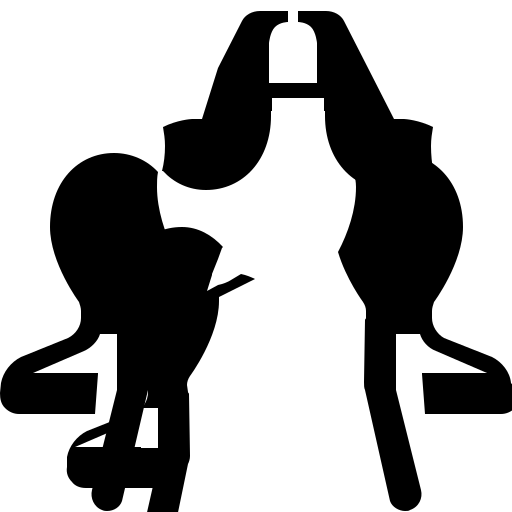}
\end{tabular*}
\caption{\textbf{Structural Coherence (SC).} Sample 5756 requests a stylized star on a blue background, and Sample 3546 requests three simple black human silhouettes. The higher-SC outputs preserve closed contours and keep distinct parts separate; the lower-SC outputs exhibit fragmentation or unintended fusion.}
\label{fig:sc_metric_semantics}
\end{figure*}

A higher SC score therefore means that the intended contours can be followed and the depicted parts form a coherent object; it does not simply reward greater detail. In Figure~\ref{fig:sc_metric_semantics}, the higher-scoring star remains one closed shape instead of fragmenting into displaced pieces. Likewise, the higher-scoring human silhouettes remain distinct rather than colliding and merging.

\subsection{Spatial Logic (SL)}
\label{app:sl_metric_semantics}

Latent Canvas defines Spatial Logic as a $[0,1]$ VLM-judge score for whether object parts have anatomically or geometrically correct relative positions~\citep{zheng2026latentcanvas}. We retain this score range when transferring the rubric to SVG.

\noindent\textbf{SL clause in the frozen joint judge prompt.}
\emph{``\texttt{latent\_spatial\_logic} $[0,1]$, adapted from Latent Canvas: are anatomy, geometry, counts, relative positions, containment, pose and placement logically correct for this exact caption?''}

This clause evaluates object and part counts, relative position, containment, pose, direction, anatomy, and geometric placement. Unlike SC, which focuses on whether shapes are structurally intact, SL asks whether the right entities appear in the right roles and locations specified by the caption.

\begin{figure*}[p]
\centering
\setlength{\tabcolsep}{1.5pt}
\begin{tabular*}{\textwidth}{@{\extracolsep{\fill}}cccc@{}}
\multicolumn{2}{c}{\small\textbf{Sample 0529}}
& \multicolumn{2}{c}{\small\textbf{Sample 1074}} \\
{\scriptsize\textbf{Higher SL}}
& {\scriptsize\textbf{Lower SL}}
& {\scriptsize\textbf{Higher SL}}
& {\scriptsize\textbf{Lower SL}} \\[-0.5mm]
\includegraphics[width=.235\textwidth]{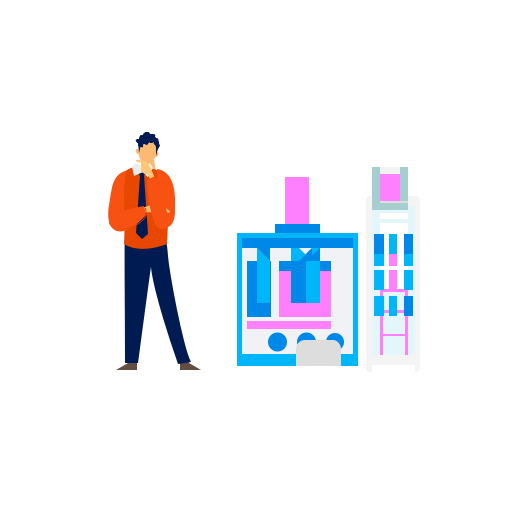}
& \includegraphics[width=.235\textwidth]{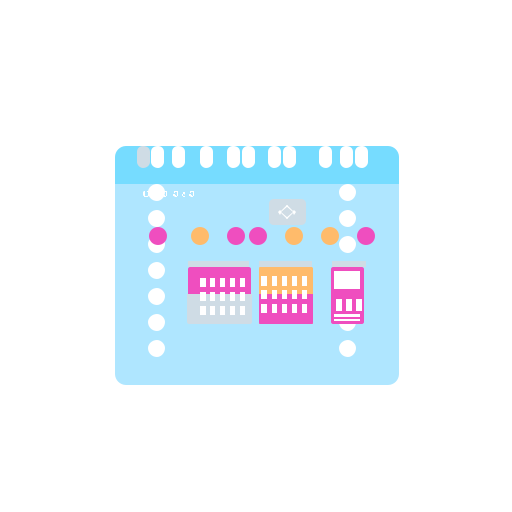}
& \includegraphics[width=.235\textwidth]{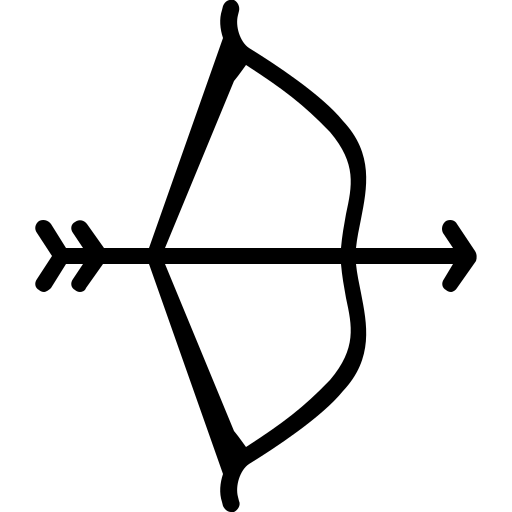}
& \includegraphics[width=.235\textwidth]{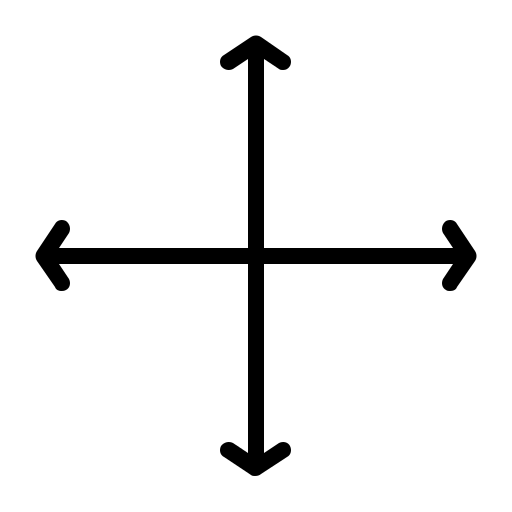}
\end{tabular*}
\caption{\textbf{Spatial Logic (SL).} Sample 0529 requests a person standing on the left, and Sample 1074 requests a line-drawn bow and arrow. The higher-SL outputs satisfy the requested placement and component relations; the lower-SL outputs change object presence, direction, or component identity.}
\label{fig:sl_metric_semantics}
\end{figure*}

A higher SL score means that the requested arrangement and component relations are represented correctly. In Figure~\ref{fig:sl_metric_semantics}, Sample 0529 tests whether the person is actually placed on the left. Sample 1074 tests whether the components form a bow and arrow rather than a clean but semantically different four-way arrow symbol.

\subsection{Character Efficiency (CE)}
\label{app:ce_metric_semantics}

Latent Canvas defines Character Efficiency as a $[0,1]$ VLM-judge score for visual cleanliness, with penalties for character spam, grid filling, and unnecessary artifacts~\citep{zheng2026latentcanvas}. We retain this score range but transfer the notion of character efficiency to the graphical elements and source structure of an SVG.

\noindent\textbf{CE clause in the frozen joint judge prompt.}
\emph{``\texttt{latent\_character\_efficiency} $[0,1]$, adapted from Latent Canvas: is the SVG clean and economical, without visual noise, repetitive path spam, duplicated marks or unnecessary artifacts? Use the supplied source statistics as evidence but do not reward an empty or under-detailed graphic.''}

This clause evaluates whether the visible structure is produced by purposeful elements rather than redundant code or over-drawing. It receives the common visual inputs together with shape, path, stratum, duplicate, and character counts. It penalizes repeated markers, unnecessary details, visual noise, and local path spam, while explicitly avoiding a trivial preference for blank or under-detailed outputs.

\begin{figure*}[p]
\centering
\setlength{\tabcolsep}{1.5pt}
\begin{tabular*}{\textwidth}{@{\extracolsep{\fill}}cccc@{}}
\multicolumn{2}{c}{\small\textbf{Sample 0653}}
& \multicolumn{2}{c}{\small\textbf{Sample 1541}} \\
{\scriptsize\textbf{Higher CE}}
& {\scriptsize\textbf{Lower CE}}
& {\scriptsize\textbf{Higher CE}}
& {\scriptsize\textbf{Lower CE}} \\[-0.5mm]
\includegraphics[width=.235\textwidth]{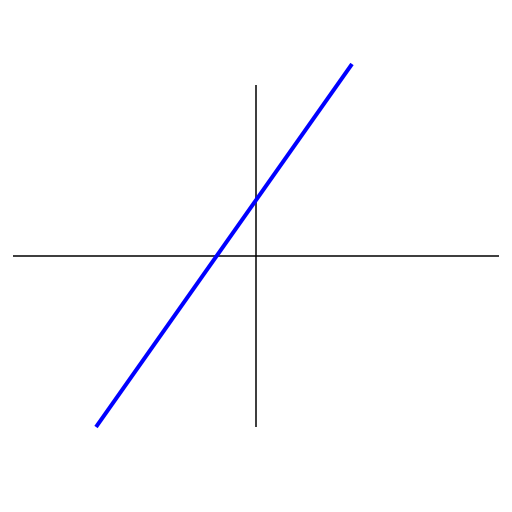}
& \includegraphics[width=.235\textwidth]{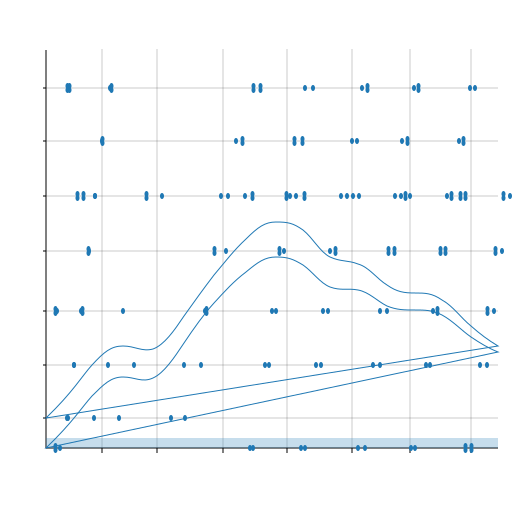}
& \includegraphics[width=.235\textwidth]{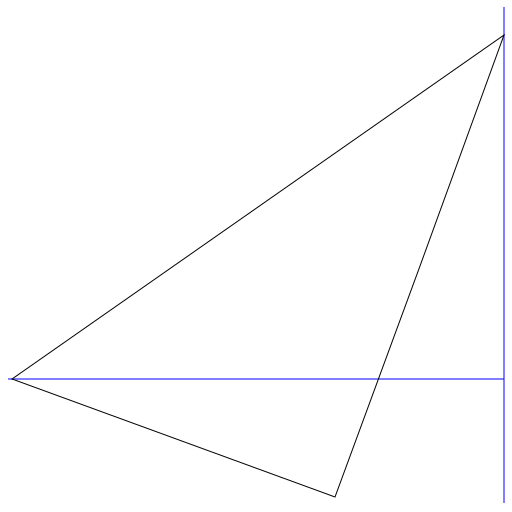}
& \includegraphics[width=.235\textwidth]{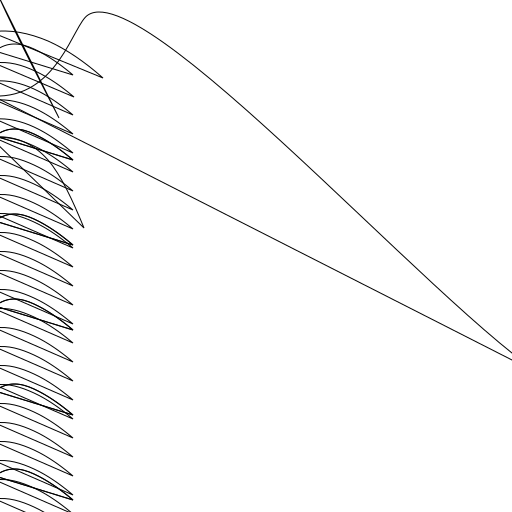}
\end{tabular*}
\caption{\textbf{Character Efficiency (CE).} Sample 0653 requests a simple graph with one blue line, and Sample 1541 requests a geometric figure made of several lines and shapes. The higher-CE outputs use a small set of purposeful elements, whereas the lower-CE outputs are dominated by unnecessary grids or repeated strokes. All four outputs are render-valid and nonempty.}
\label{fig:ce_metric_semantics}
\end{figure*}

A higher CE score therefore indicates that most generated elements make a useful contribution to the final image. In Figure~\ref{fig:ce_metric_semantics}, the higher-scoring outputs use three graphic elements in each pair. Their lower-scoring counterparts use 152 and 167 elements, respectively, to create dense grids or repeated hatching that the prompts do not require.

\subsection{LaDe Cross-Layer Consistency}
\label{app:lade_metric_semantics}

LaDe defines a $1$--$5$ VLM-judge rubric for layered designs that includes cross-layer consistency among its evaluation stages~\citep{lungustan2026lade}. We retain this score range while transferring that stage to SVG pseudo-layers.

\noindent\textbf{LaDe clause in the frozen joint judge prompt.}
\emph{``\texttt{lade\_cross\_layer\_consistency} $[1,5]$, adapted from LaDe: across the ordered pseudo-layers, assess color harmony, style, perspective, scale, occlusion/depth, foreground/background relations and whether components form one coherent composite. 1 is severe conflict; 5 is fully consistent.''}

This clause measures whether separately drawn components combine into one coherent scene. The original LaDe protocol evaluates semantic RGBA layers as part of a broader layered-design rubric and uses different judges. Our flat-SVG evaluation instead groups contiguous drawables into ordered pseudo-layers, so the resulting score is a transfer proxy rather than the native LaDe score.

\clearpage
\begin{figure*}[t]
\centering
\setlength{\tabcolsep}{1.5pt}
\begin{tabular*}{\textwidth}{@{\extracolsep{\fill}}cccc@{}}
\multicolumn{2}{c}{\small\textbf{Sample 3293}}
& \multicolumn{2}{c}{\small\textbf{Sample 2693}} \\
{\scriptsize\textbf{Higher LaDe}}
& {\scriptsize\textbf{Lower LaDe}}
& {\scriptsize\textbf{Higher LaDe}}
& {\scriptsize\textbf{Lower LaDe}} \\[-0.5mm]
\includegraphics[width=.235\textwidth]{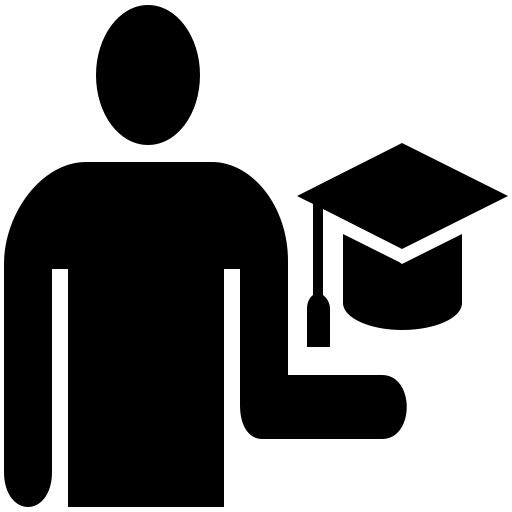}
& \includegraphics[width=.235\textwidth]{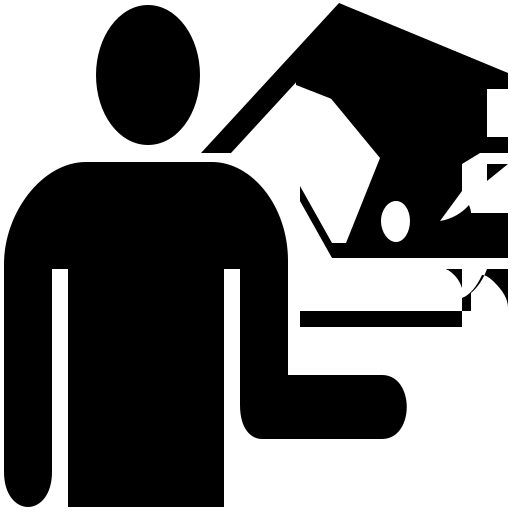}
& \includegraphics[width=.235\textwidth]{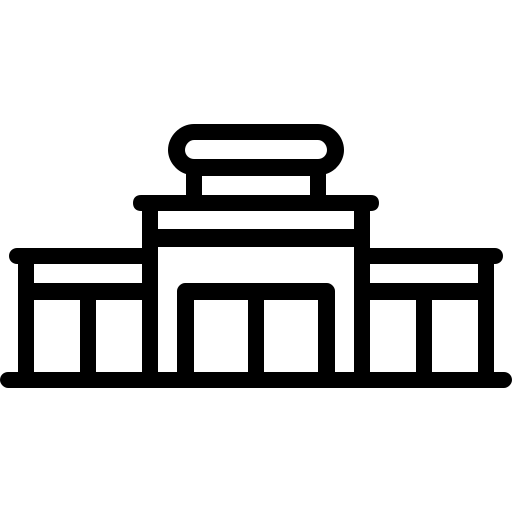}
& \includegraphics[width=.235\textwidth]{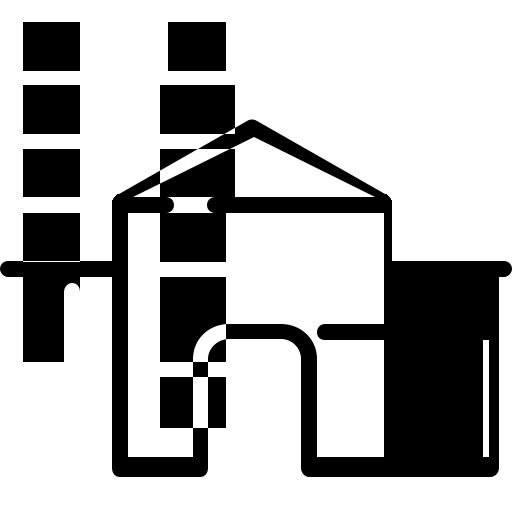}
\end{tabular*}
\caption{\textbf{LaDe Cross-Layer Consistency.} Sample 3293 requests a person wearing a graduation cap and gown, and Sample 2693 requests a two-dimensional building. The higher-LaDe outputs combine their parts at compatible scales and with clearer occlusion order; the lower-LaDe outputs contain conflicting overlays and front-to-back relations.}
\label{fig:lade_metric_semantics}
\end{figure*}

A higher transferred LaDe score means that components agree in scale and appearance and follow a sensible front-to-back order. In Figure~\ref{fig:lade_metric_semantics}, the higher-scoring graduation figure aligns its cap, head, gown, and arms, while the lower-scoring output overlays them at incompatible scales. The building pair similarly reveals differences in roof, facade, entrance, and foreground occlusion.

\subsection{\texorpdfstring{Local Connectivity Index (LCI$_{9\times9}$)}{Local Connectivity Index (LCI 9x9)}}
\label{app:lci_metric_semantics}

To measure local contour connectivity, LCI$_{9\times9}$ adapts the grid-based line-connectivity measure of~\citet{feng2025mural}. Divide the shared edge map into a $9\times9$ grid and let $\mathcal G_+$ be the set of nonempty tiles. For each $g\in\mathcal G_+$, let $n_g$ be its number of edge pixels and let $c_g$ be the number of pixels in its largest 8-connected component. For a nonempty edge map, we compute
\begin{equation}
    \mathrm{LCI}_{9\times9}
    =\frac{1}{|\mathcal G_+|}
    \sum_{g\in\mathcal G_+}\frac{c_g}{n_g}
    \in[0,1].
    \label{eq:lci_metric_definition}
\end{equation}
Each tile contributes the fraction of its edges belonging to its dominant connected component, and the final score averages these fractions over occupied tiles.

\begin{figure*}[t]
\centering
\setlength{\tabcolsep}{1.5pt}
\begin{tabular*}{\textwidth}{@{\extracolsep{\fill}}cccc@{}}
\multicolumn{2}{c}{\small\textbf{Sample 0543}}
& \multicolumn{2}{c}{\small\textbf{Sample 1245}} \\
{\scriptsize\textbf{Higher LCI}}
& {\scriptsize\textbf{Lower LCI}}
& {\scriptsize\textbf{Higher LCI}}
& {\scriptsize\textbf{Lower LCI}} \\[-0.5mm]
\includegraphics[width=.235\textwidth]{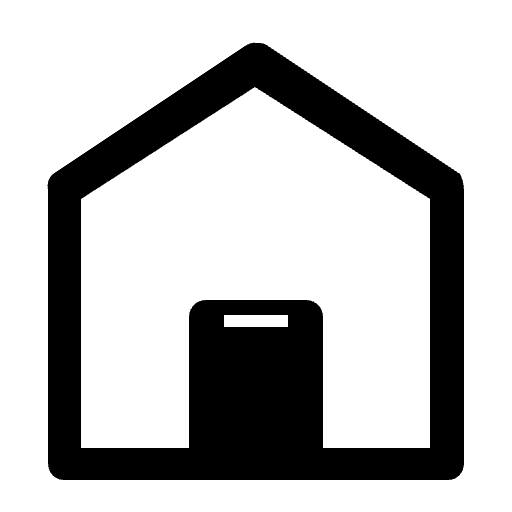}
& \includegraphics[width=.235\textwidth]{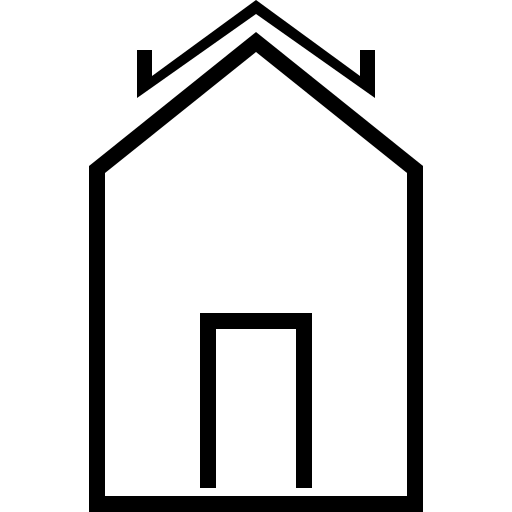}
& \includegraphics[width=.235\textwidth]{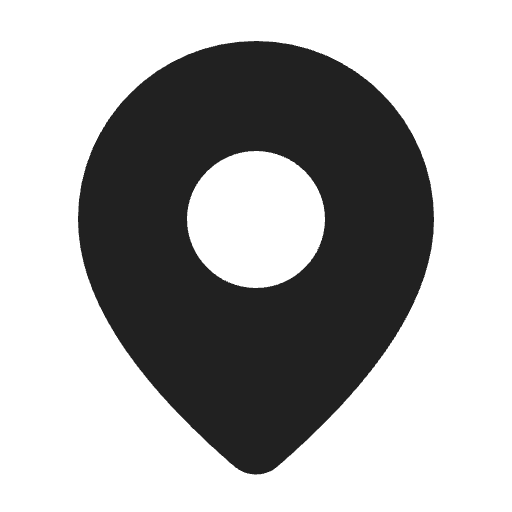}
& \includegraphics[width=.235\textwidth]{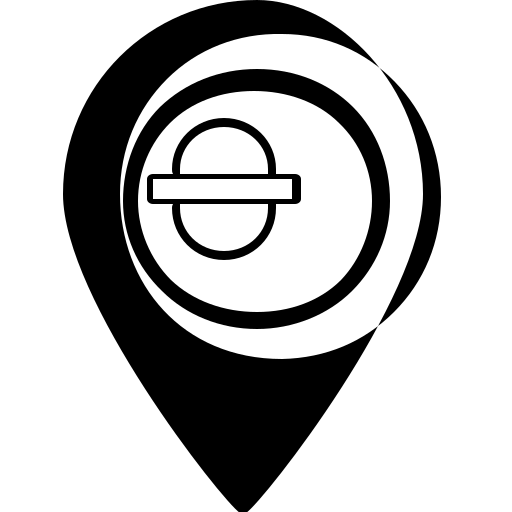}
\end{tabular*}
\caption{\textbf{Local Connectivity Index (LCI$_{9\times9}$).} Sample 0543 requests a simple house outline, and Sample 1245 requests a minimalist map marker. The higher-LCI outputs contain one dominant connected contour within most occupied grid cells; the lower-LCI outputs contain detached or competing local edge components.}
\label{fig:lci_metric_semantics}
\end{figure*}

A higher LCI$_{9\times9}$ score means that local regions are usually dominated by one connected contour rather than several detached or competing fragments. It does not directly measure prompt agreement, detail count, or how widely the drawing covers the canvas. Figure~\ref{fig:lci_metric_semantics} illustrates this distinction with a continuous house outline versus detached roof and doorway fragments, and with a simple map-marker contour versus several nearby boundaries.

\newpage
\subsection{CVIU Complexity (CVIU-C)}
\label{app:cviu_metric_semantics}

To measure whether an SVG combines regular local edge structure with nondegenerate spatial coverage, CVIU-C adapts the reference-free Statistical Complexity Measure of~\citet{gimenez2014edge}. Let $\mathcal P$ be the set of edge pixels in the shared edge map, $N_p$ the vectorized $7\times7$ neighborhood around edge pixel $p$, and $\mathcal T$ the bank of 140 local line patterns from the original measure. Let $\mathcal U$ denote the uniform distribution over canvas locations, and let $D_{2\mathrm D}(\mathcal P,\mathcal U)\in[0,1]$ be the maximum two-dimensional empirical-CDF discrepancy between the observed edge locations and $\mathcal U$. We compute
\begin{equation}
\begin{aligned}
    E
    &=\frac{1}{|\mathcal P|}\sum_{p\in\mathcal P}
    \max_{\tau\in\mathcal T}
    \frac{\langle N_p,\tau\rangle}
    {\|N_p\|_2\|\tau\|_2},\\
    H&=1-D_{2\mathrm D}(\mathcal P,\mathcal U),
    \qquad
    \mathrm{CVIU\text{-}C}=E\,H\in[0,1].
\end{aligned}
    \label{eq:cviu_metric_definition}
\end{equation}
The equilibrium term $E$ is the average best match to a regular local line pattern. The information term $H$ decreases when edges collapse onto one axis or a small region.

\begin{figure*}[t]
\centering
\setlength{\tabcolsep}{1.5pt}
\begin{tabular*}{\textwidth}{@{\extracolsep{\fill}}cccc@{}}
\multicolumn{2}{c}{\small\textbf{Sample 0557}}
& \multicolumn{2}{c}{\small\textbf{Sample 5291}} \\
{\scriptsize\textbf{Higher CVIU-C}}
& {\scriptsize\textbf{Lower CVIU-C}}
& {\scriptsize\textbf{Higher CVIU-C}}
& {\scriptsize\textbf{Lower CVIU-C}} \\[-0.5mm]
\includegraphics[width=.235\textwidth]{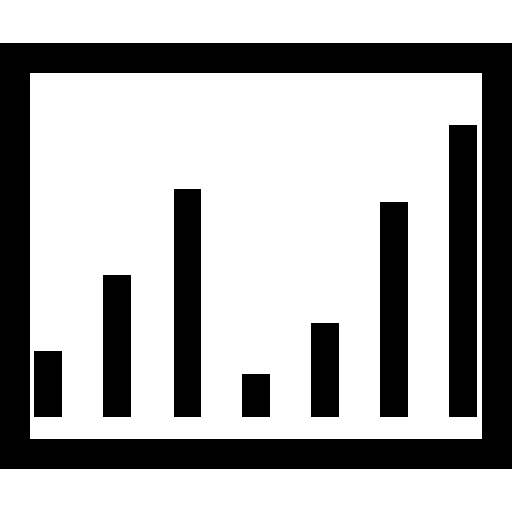}
& \includegraphics[width=.235\textwidth]{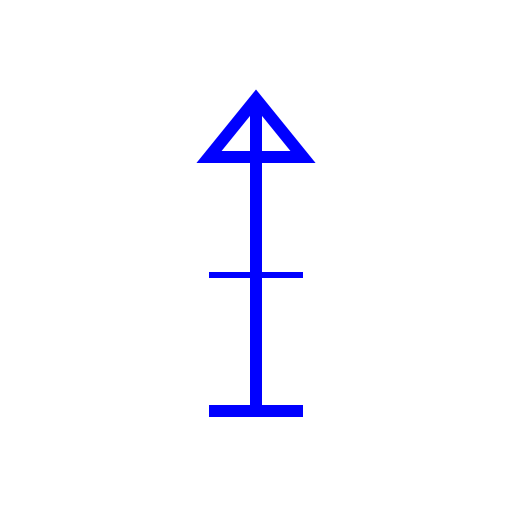}
& \includegraphics[width=.235\textwidth]{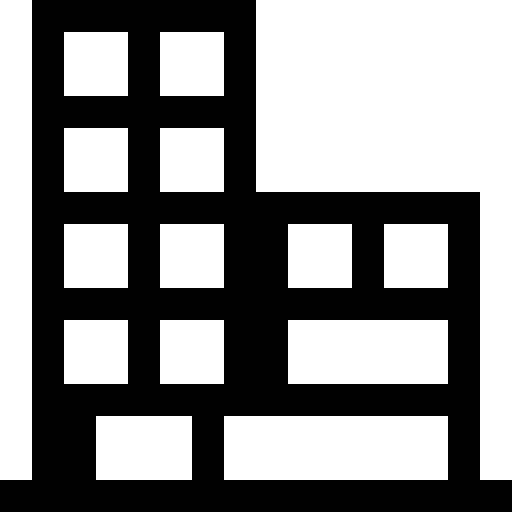}
& \includegraphics[width=.235\textwidth]{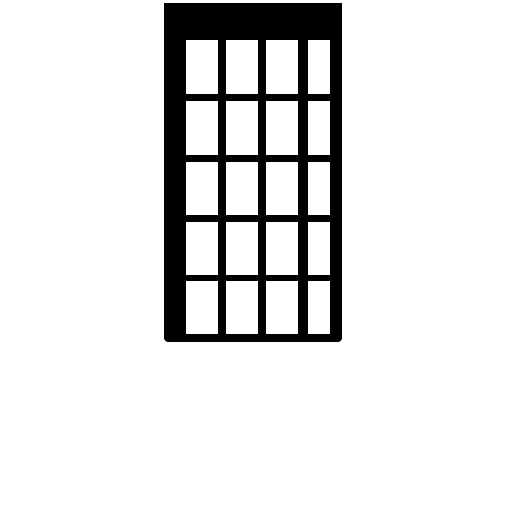}
\end{tabular*}
\caption{\textbf{CVIU Complexity (CVIU-C).} Sample 0557 requests a minimalist vertical bar chart, and Sample 5291 requests a two-dimensional building. The higher-CVIU-C outputs distribute regular edge motifs across both canvas axes; the lower-CVIU-C outputs concentrate most edge information along one axis or within a narrow region.}
\label{fig:cviu_metric_semantics}
\end{figure*}

A higher CVIU-C score therefore requires both clean local edge motifs and sufficiently rich two-dimensional coverage. This differs from LCI$_{9\times9}$: a small or one-axis drawing may be locally connected yet still receive low CVIU-C because its $H$ term is small. Figure~\ref{fig:cviu_metric_semantics} shows this difference through a distributed bar chart versus a concentrated arrow-like axis, and a broad building facade versus a narrow vertical edge pattern.

\FloatBarrier

\FloatBarrier
\end{document}